\documentclass{article} 
\usepackage{iclr2027_conference,times}

\usepackage{amsmath,amsfonts,bm}

\def\eqref#1{equation~\ref{#1}}

\def\1{\bm{1}}

\DeclareMathAlphabet{\mathsfit}{\encodingdefault}{\sfdefault}{m}{sl}
\SetMathAlphabet{\mathsfit}{bold}{\encodingdefault}{\sfdefault}{bx}{n}

\usepackage{hyperref}
\usepackage{url}
\usepackage{graphicx}
\usepackage{amssymb}
\usepackage{booktabs}
\usepackage{longtable}
\usepackage{multirow}
\usepackage{subcaption}

\definecolor{tgtest}{HTML}{2F76B0}
\definecolor{tgrules}{HTML}{C2571A}
\definecolor{tgband}{HTML}{F5E4DA}

\definecolor{tgeasy}{HTML}{EAF2E8}
\definecolor{tgmed}{HTML}{FBF1DF}
\definecolor{tghard}{HTML}{F8E7E4}
\definecolor{tgshort}{HTML}{ECECEC}
\definecolor{tglong}{HTML}{E7E0F0}
\definecolor{tglvlone}{HTML}{B5D4EA}
\definecolor{tglvlseven}{HTML}{0F3F70}

\newcommand{\swline}[1]{%
  \textcolor{#1}{\rule[0.32ex]{11pt}{1.5pt}}}
\newcommand{\swdash}[1]{%
  \textcolor{#1}{\rule[0.32ex]{4pt}{1.5pt}\hspace{2pt}%
                 \rule[0.32ex]{4pt}{1.5pt}}}
\newcommand{\swband}[2]{%
  {\setlength{\fboxsep}{0pt}%
   \fcolorbox{#1}{#2}{\rule{0pt}{1.15ex}\rule{11pt}{0pt}}}}
\newcommand{\swgrey}{%
  \textcolor{black!30}{\rule[0.32ex]{3pt}{1.2pt}\hspace{2pt}%
                       \rule[0.32ex]{3pt}{1.2pt}}}

\title{What Do Current Systematic Generalization Tasks Miss? A Reasoning-Centered Analysis}

\author{Chengwen Qi$^{1}$ \quad Deheng Ye$^{2}$ \quad Yatao Bian$^{1}$\thanks{Corresponding author}\\
$^1$National University of Singapore \quad 
$^2$Nanyang Technological University\quad \\
\texttt{qichengwen@u.nus.edu},\ \ \texttt{ydyl1991@gmail.com} \\
\texttt{ybian@nus.edu.sg} \\
}

\iclrfinalcopy 
\begin{document}

\maketitle

\begin{abstract}
Systematic generalization, the ability to solve novel problems by recombining known atomic elements, is central to human intelligence but difficult to study rigorously under controlled settings. 
Existing studies therefore rely on simplifications such as elemental composition, productivity-based tests, and action-explicit goals, which make systematic generalization easier to study but omit some essential aspects of this capability.
To characterize what these simplifications miss, we adopt a reasoning-centered lens and introduce TranSGrid, a testbed that brings deductive, inductive, and abductive reasoning together within a unified task. 
Experiments with seven Transformer models on 4,800 TranSGrid instances show that all models perform much worse on TranSGrid than on a held-out test set: the largest model solves 79.6\% of the test set, but only 55.3\% of TranSGrid and 15.8\% of the hardest subset. 
The gap remains within the training length range, showing that productivity alone is not sufficient to evaluate systematic generalization. 
Additionally, we reintroduce the other two simplifications into TranSGrid: one variant limits interactions among action effects to approximate elemental composition (reducing the inductive demand); the other makes goals action-explicit (reducing the abductive one). 
In both, solve rates return to roughly the test set level, showing that either simplification alone is enough to reduce TranSGrid to an ordinary held-out test set. 
Together, our results show that existing tasks reduce either or both of the inductive and abductive demands of systematic generalization, and that comprehensively measuring this capability requires a task that involves all three forms of reasoning. \footnote{Code available at: \url{https://github.com/BlueWhaleLab/TranSGrid}}
\end{abstract}

\section{Introduction}
\label{sec:intro}

Systematic generalization, which refers to the ability to systematically combine learned actions or skills to address novel problems, is a fundamental aspect of human intelligence~\citep{lake2018generalization, lake2023human}. 
This capability is essential across diverse domains, including grounded navigation~\citep{ruis2020benchmark, sikarwar2022can, spilsbury-etal-2024-generating, chen2026rule}, semantic parsing~\citep{kim2020cogs, wu2023recogs, jabbar-etal-2025-distinguishing}, text-to-image generation~\citep{han2025progressive, huang2025t2i, dat2025vsc}, facial recognition~\citep{rotshtein2007role, leong2023holistic}, AI safety~\citep{addepalli2025does, chen2026sage}, and AI for Science~\citep{ji2023drugood, chen2025hierarchical, li2026speak, li2026llms}. 
Across these domains, success depends not merely on mastering individual components, but also on understanding how they interact and recombining them appropriately under novel goals, contexts, and constraints. 

A rigorous study of systematic generalization therefore requires 
(1) a \emph{clearly defined space of atomic actions},  
(2) \emph{meaningful, interaction-rich compositions},  
(3) \emph{scalable generation} of instances that probe models' systematicity, and 
(4) \emph{automatic evaluation} of multiple valid solutions. 
In practice, however, satisfying all these requirements within a single task is difficult, leading existing studies to rely on simplifications in task design. 
One common simplification is elemental composition\footnote{This terminology is adapted from studies of human and animal learning~\citep{devaud2015neural, duncan2018more}: elemental learning predicts a combination's outcome from its parts alone, whereas configural learning also uses predictive information from how those parts are combined.}, in which interactions among atomic elements are limited~\citep{lake2018generalization, lake2023human}.
A second is to assess systematicity primarily through productivity, typically by testing whether models extrapolate to sequences longer than those observed during training~\citep{hupkes2020compositionality, fu2026reinforcement}. 
A third is to use action-explicit goals, where the input already specifies the actions and their order, largely reducing the task to interpreting a given composition rather than finding a valid action sequence within a vast combinatorial space~\citep{wu2023recogs, li2023a, mondorf2026compositionalarc}. 
Although these simplifications make systematic generalization easier to study, they remove some of its core challenges and thus provide an incomplete picture of the capability.

To examine what these simplifications ignore, \textbf{we employ a reasoning-centered lens to study systematic generalization} and introduce \textbf{TranS}formation on a \textbf{Grid} (\textbf{TranSGrid}), a controllable and flexible testbed. 
Inspired by cognitive science research highlighting the interplay of deduction, induction, and abduction in human intelligence~\citep{peirce1934collected,shank1998extraordinary,chemero2026abduction}, TranSGrid is designed to operationalize this interplay within a unified task. 
Specifically, TranSGrid requires a model to transform an initial board into a target board by generating an action sequence drawn from ten atomic operations on rows, columns, or local $2\times2$ blocks.
The task involves deductive reasoning to derive and track the board state after each action.
Its inductive challenge lies in discovering reusable composition rules from observed examples of how action effects combine, cancel, or obscure one another, and applying these rules to novel instances. 
Abductive reasoning involves inferring a suitable action sequence from the desired outcome. 
To accommodate multiple valid solutions, TranSGrid evaluates the correctness of each generated sequence by executing it on the initial board and checking whether the resulting board matches the target. 
Any sequence that produces the target board is considered correct; it need neither be a shortest solution nor match the reference sequence. 

Experiments with seven Transformer models ranging from 0.96M to 88.43M parameters on 4,800 TranSGrid instances show that all models perform substantially worse on TranSGrid than on a held-out Test set. 
Under greedy decoding, the largest model's solve rate falls from 79.63\% on Test to 55.31\% on TranSGrid overall, reaching only 15.75\% on TranSGrid$_{\mathrm{Hard}}$. 
Crucially, this gap persists even when reference answer lengths fall within the training range, showing that productivity alone is insufficient to evaluate systematic generalization. 
We then reintroduce the other two simplifications through controlled variants. 
TranSGrid (Decoupled) approximates elemental composition by limiting interactions among action effects, thus reducing the inductive demand of inferring reusable interaction rules from examples.
TranSGrid (SCAN) makes goals largely action-explicit by providing all but the final gold action, reducing the abductive demand of inferring a suitable action sequence from the desired outcome. 
Solve rates in both variants recover to roughly the held-out test set level, indicating that either simplification substantially reduces the challenge posed by TranSGrid. 
Taken together, these findings show that existing tasks overlook essential inductive and abductive reasoning demands of systematic generalization, and that a comprehensive evaluation of this capability requires tasks that involve all three forms of reasoning: deduction, induction, and abduction.

\section{Related Work}
\label{sec:related_work}

Existing benchmarks for systematic generalization can be broadly grouped into three categories. 
First, action or solution-explicit tasks, such as SCAN~\citep{lake2018generalization}, PCFG~\citep{hupkes2020compositionality}, COGS~\citep{kim2020cogs}, and HINT~\citep{li2023a}, provide the action sequence directly in the input. 
Models therefore need only to parse or execute the given composition rather than infer one from a goal. 
Second, grounded navigation benchmarks, including gSCAN~\citep{ruis2020benchmark} and ReaSCAN~\citep{wu2021reascan}, require models to generate an action sequence in a structured world. 
However, once the target is identified, the route can usually be derived directly from the world, so these benchmarks focus more on language grounding and target localization than on abductive reasoning. 
Action-action interactions are also present, but are usually simple.
Across these two lines, many evaluations emphasize productivity by testing sequences longer or deeper than those seen during training. 
Finally, rule-changing tasks such as \textit{Baba Is You} provide richer action interactions by requiring agents to create, break, and combine rules to reach a goal~\citep{cloos2024baba}. 
However, its visual input and various distractors make it difficult to determine whether failures result from weak systematic generalization or poor scene understanding. 
TranSGrid addresses these limitations by requiring models to infer action sequences from goals and reason about interacting action effects in a simple, controllable setting, allowing systematic generalization to be studied more reliably and with fewer confounding factors.

\section{A Grid-Transformation Task for Systematic Generalization}
\label{sec:preliminary}

\subsection{Challenges in Measuring Systematic Generalization}
\label{subsec:preliminary_difficulty}

Although systematic generalization is essential across many domains, constructing a rigorous and scalable benchmark remains difficult. 
Such a benchmark needs to combine a well-defined atomic action space, automatic generation of meaningful compositions, scalable data construction, and automatic evaluation of multiple valid solutions. 
Yet most existing task settings satisfy only a subset of these requirements. 
For example, there is no precise definition of atomic actions in code generation; meaningful action interactions in spreadsheet editing often require substantial human effort; and tasks such as \textit{Baba Is You} introduce perceptual confounds unrelated to systematic generalization. 
Beyond these construction challenges, there is also no consensus on how to measure systematic generalization: most criteria are task-specific, with productivity being the only one widely shared across tasks. 
However, productivity is not a necessary condition for systematic generalization, as generalization within the training length range can also be systematic. 
Together, these construction and evaluation challenges help explain why prior work relies on simplifications such as elemental composition, productivity-based evaluation, and action-explicit goals. 
They also motivate a testbed that addresses these construction challenges and the limitations of productivity-based evaluation. 

\subsection{Design of the TranSGrid Task}
\label{subsec:preliminary_task}

\begin{figure}[htb]
\centering
\begin{subfigure}{0.45\linewidth}
  \centering
  \includegraphics[width=\linewidth]{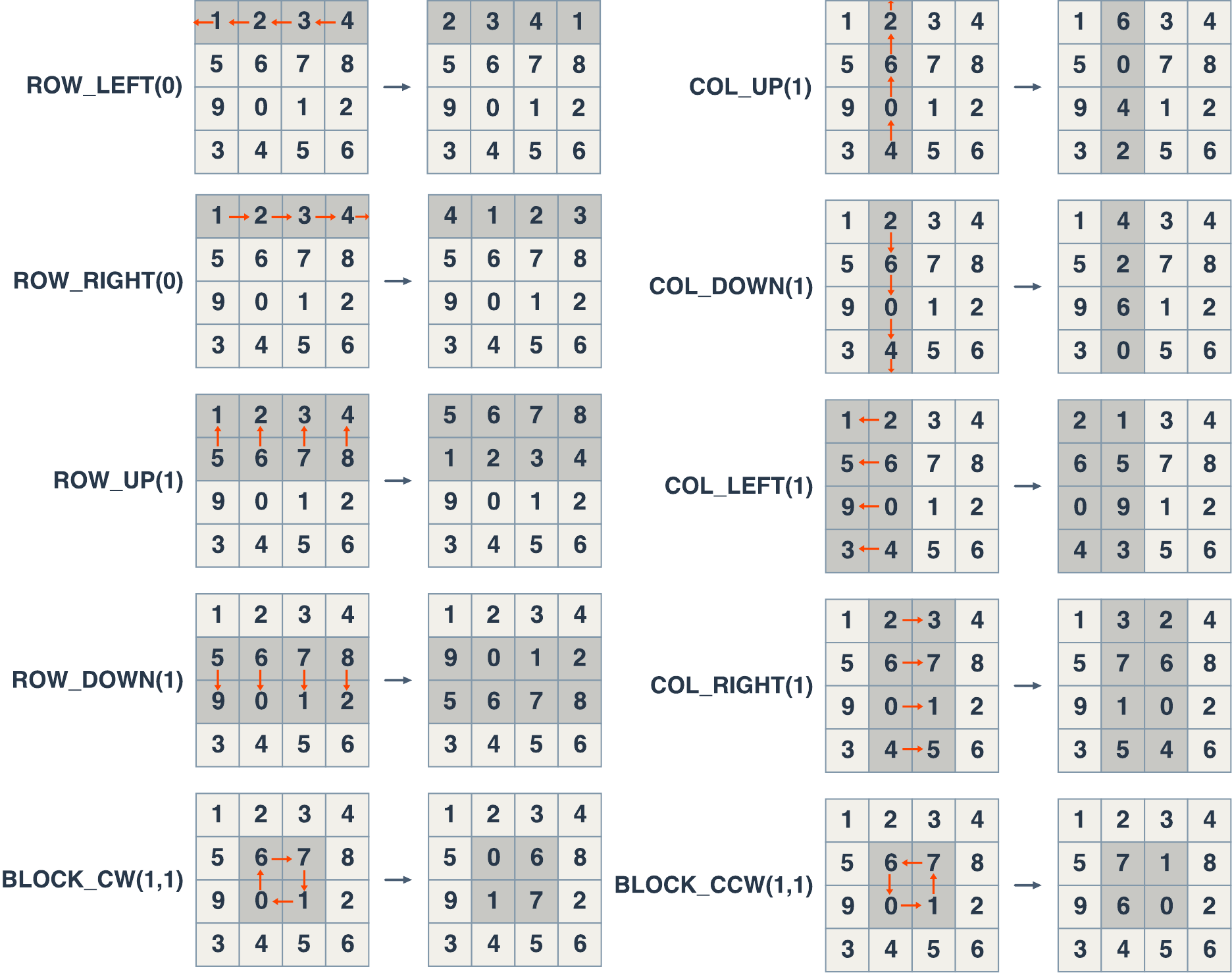}
  \caption{}
  \label{fig:atomic_action}
\end{subfigure}
\hfill
\begin{subfigure}{0.50\linewidth}
  \centering
  \includegraphics[width=\linewidth]{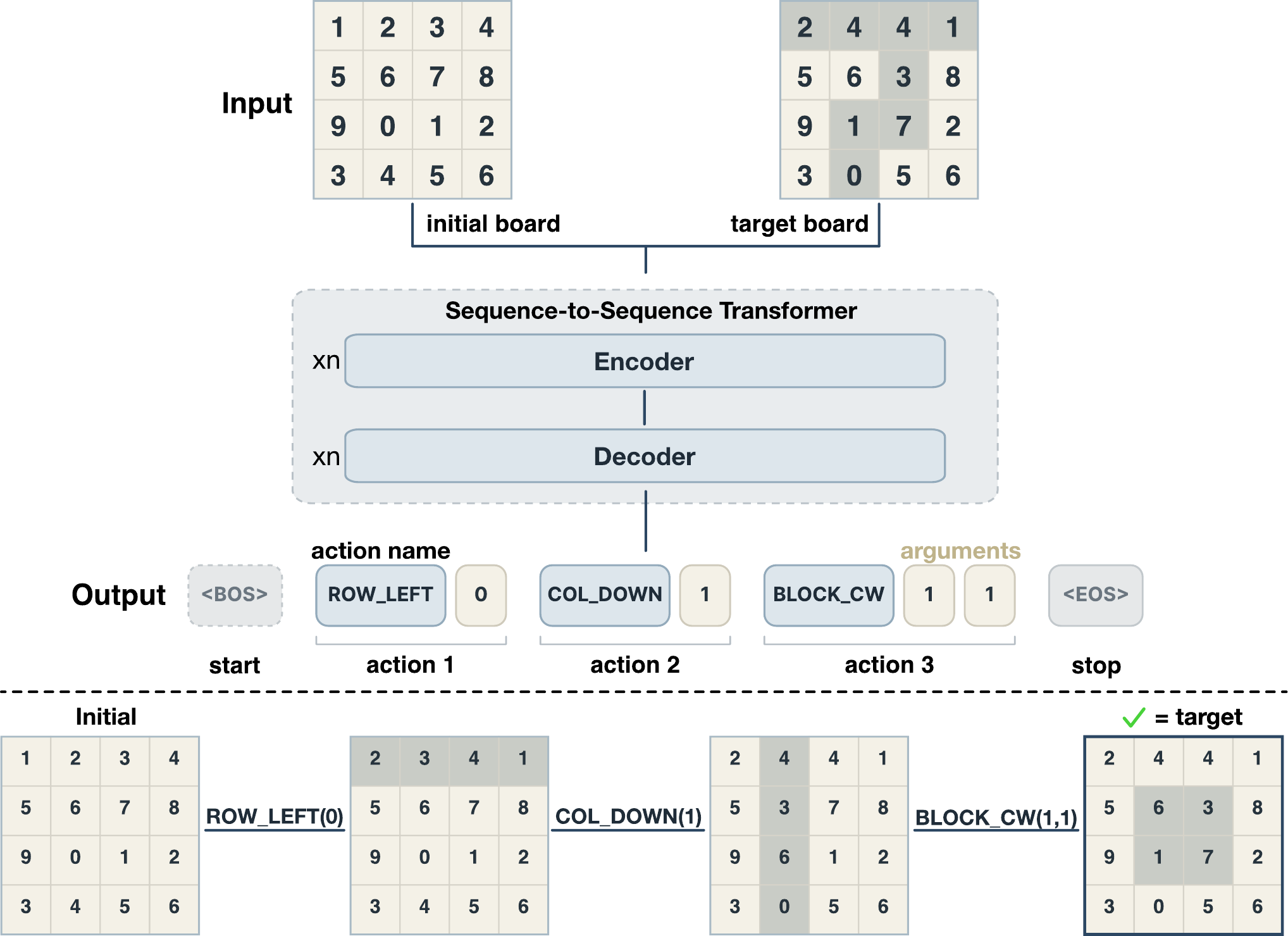}
  \caption{}
  \label{fig:task}
\end{subfigure}
\caption{The proposed TranSGrid task with its atomic actions. (a) The ten types of atomic actions defined for grid manipulation, including row-wise shifts, column-wise shifts, and rotations of local blocks. Red arrows indicate the direction of each transformation. (b) Illustration of the TranSGrid task. Given an initial board and a target board, the models need to generate an action sequence that, when executed sequentially, can transform the initial board into the target board.}
\label{fig:task_description}
\end{figure}

To address these challenges, we introduce \textbf{TranS}formation on a \textbf{Grid} (\textbf{TranSGrid}), a controllable and flexible grid-transformation testbed for systematic generalization.
Figure~\ref{fig:task_description} provides an overview of the task, with its atomic action space and task formulation detailed below. 

\textbf{Atomic action space }
As illustrated in Figure~\ref{fig:atomic_action}, TranSGrid defines ten types of atomic actions. 
\texttt{ROW\_LEFT} and \texttt{ROW\_RIGHT} cyclically shift the elements of a selected row, while \texttt{COL\_UP} and \texttt{COL\_DOWN} perform the analogous operation on a column. 
\texttt{ROW\_UP} and \texttt{ROW\_DOWN} move a selected row by swapping it with an adjacent row, and \texttt{COL\_LEFT} and \texttt{COL\_RIGHT} similarly move a selected column. 
Finally, \texttt{BLOCK\_CW} and \texttt{BLOCK\_CCW} rotate a $2\times2$ subgrid $90^\circ$ in either direction.

By design, each atomic action modifies multiple cells at once, operating on an entire row, column, or $2\times2$ block. 
Because actions may affect overlapping regions and are executed sequentially, their effects are both state-dependent and interdependent: a later action may reinforce, overwrite, or reverse the effect of an earlier one. 
Solving TranSGrid therefore requires models to track the evolving board and reason about both action-action and action-environment interactions, providing a controlled, interaction-rich setting for studying systematic generalization. 

\textbf{Task formulation }
A TranSGrid instance consists of an initial board and a target board, both of size $N\times N$, as shown in Figure~\ref{fig:task}.\footnote{All experiments in this paper use $6\times6$ boards, but for visual clarity, we use $4\times4$ boards in the figures.}
Let
\[
B^{(0)}, B^\star \in \mathcal{V}^{N\times N},
\quad
\mathcal{V}=\{0,1,\ldots,9\},
\]
denote the initial and target boards, respectively, where $B^{(0)}_{i,j}$ is the value in row $i$ and column $j$ of the initial board. 
TranSGrid provides ten atomic operation types,
\[
\mathcal{A}=\{\alpha_1,\ldots,\alpha_{10}\},
\]
each of which performs a deterministic transformation on the board, as illustrated in Figure~\ref{fig:atomic_action}. 
Let
\[
T:\mathcal{A}\times\mathcal{V}^{N\times N}
\rightarrow\mathcal{V}^{N\times N}
\]
denote the transition function, such that $T(a,B)$ is the board obtained by applying action $a$ to board $B$.
Composition in TranSGrid is sequential. An action sequence of length $L$ is defined as
\[
\pi=(a_1,\ldots,a_L)\in\mathcal{A}^L.
\]
Executing $\pi$ from $B^{(0)}$ yields
\[
B^{(t)}=T\bigl(a_t,B^{(t-1)}\bigr),
\qquad t=1,\ldots,L.
\]
The final board after executing $\pi$ is therefore $B^{(L)}$. 
Given an instance $(B^{(0)},B^\star)$, the model is required to generate a sequence $\pi$ such that $B^{(L)}=B^\star$. 
Any sequence satisfying this condition is considered a valid solution; it need neither be a shortest solution nor match the reference sequence.

\subsection{Controllability, Flexibility, and Scalability of TranSGrid}
\label{subsec:transgrid_property}

TranSGrid provides explicit control over task properties through adjustable parameters, including grid size $N$, digit distribution, reference action-sequence length, and composition patterns. 
These parameters allow task structure and difficulty to be varied systematically. 
Its flexible design can also reproduce two common simplifications in existing tasks: elemental composition and action-explicit goals (see Figures~\ref{fig:transgrid_decoupled} and~\ref{fig:transgrid_scan}, respectively, for further details).

Moreover, TranSGrid provides a vast combinatorial task space. 
For example, with $N=6$ and ten possible cell values, there are $10^{36}$ possible initial boards. 
Because the atomic actions can be applied to different rows, columns, or blocks, a $6\times6$ board admits
\[
4\cdot6+2\cdot6+2\cdot5^2=86
\]
distinct board transformations.\footnote{The ten types of atomic actions can be applied in $8\cdot6+2\cdot5^2=98$ ways. Among them, \texttt{ROW\_UP}/\texttt{ROW\_DOWN} and \texttt{COL\_LEFT}/\texttt{COL\_RIGHT} form 12 equivalent pairs, leaving 86 distinct transformations.}
Restricting the reference action-sequence length to the range 1--9 therefore yields approximately
\[
10^{36}\times\sum_{k=1}^{9}86^k
\approx10^{36}\times2.60\times10^{17}
\approx2.60\times10^{53}
\]
possible pairs of initial boards and reference action sequences. 
Although different action sequences may produce the same target board from a given initial board when their effects overlap or cancel, the task space of TranSGrid remains extremely large. 
Instances can be generated automatically by sampling an initial board and executing a reference action sequence to obtain the target board, guaranteeing a valid solution by construction. 
Together with its controllability and flexibility, this scale makes TranSGrid an ideal platform for developing new methods to improve models' systematic generalization performance and investigating the mechanisms underlying this capability.

\section{Experimental Results on TranSGrid}
\label{sec:transgrid}

\subsection{The three reasoning forms in TranSGrid}
\label{subsec:tg_reasoning_forms}

TranSGrid is designed to jointly engage deductive, inductive, and abductive reasoning within a unified task. 
Although each form corresponds to a distinct aspect of the task, the associated reasoning demands are inherently interdependent.
We next describe how these forms manifest in TranSGrid and introduce the corresponding metrics.

\textbf{Deductive Reasoning }
Deductive reasoning in TranSGrid involves inferring the next board state from the current state and action. 
Since each additional action demands one more step of forward inference over the board state, we use the reference sequence length as a proxy for deductive difficulty, with longer sequences indicating greater deductive load.
%
%

\textbf{Inductive Reasoning }
The atomic actions in TranSGrid operate on multiple cells. 
However, when combined in specific ways, their intermediate effects can cancel out, leaving only a few cells permuted. 
The inductive challenge lies in discovering reusable composition rules from observed examples of such interactions and applying these rules to new instances. 
We refer to action combinations that realize these localized permutations as induction rules and manually define 12 such rules for constructing TranSGrid instances (see Appendix~\ref{appendix:induction_rules} for details).
Accordingly, we use the number of induction rules instantiated in the reference sequence as a proxy for inductive difficulty, with larger counts representing higher inductive load.

\textbf{Abductive Reasoning }
The task formulation of TranSGrid inherently involves abductive reasoning, as the model needs to infer a plausible sequence of actions from the observed initial and target boards. 
When actions overwrite or cancel one another's effects, the resulting boards can provide fewer observable cues about the actions involved. 
To quantify this concealment, we let $D$ denote the number of cells that differ between the initial and target boards.
We then map $D$ to an effective action length $L_{\mathrm{eff}}$, which estimates how many random actions would typically produce the same amount of observable change (see Appendix~\ref{appendix:effective_length}). 
Given the reference sequence length $L$, we define
$H_{\mathrm{act}}=L-L_{\mathrm{eff}}$
as a proxy for action concealment: larger values indicate that the reference sequence contains more actions than its observable effect would typically suggest.
We then normalize this proxy to obtain an abductive score:
\[
A_{\mathrm{abd}}=\sigma\left(\frac{H_{\mathrm{act}}-6}{2}\right),
\]
where $\sigma(x)=1/(1+e^{-x})$.
Using the maximum reference sequence length $L_{\max}=12$, we set the sigmoid midpoint to $L_{\max}/2=6$ and its scale to $L_{\max}/6=2$. 
This calibration gives $A_{\mathrm{abd}}=0.5$ at $H_{\mathrm{act}}=6$, with scores of approximately 0.05 and 0.95 at $H_{\mathrm{act}}=$ 0 and 12, respectively. 
For analysis, we discretize $A_{\mathrm{abd}}$ into five equal-width levels (1--5), defined by the intervals $[0,0.2), [0.2,0.4), \ldots, [0.8,1)$, with higher levels indicating greater abductive load.

\subsection{Experimental Setup}
\label{subsec:tg_setup}

\textbf{Model }
Following prior work on systematic generalization~\citep{kim2020cogs, lake2023human, li2023a, kumon2025analyzing}, we train standard encoder-decoder Transformers to map an initial--target board pair to an action sequence. 
The encoder jointly processes both boards, while the decoder autoregressively generates action-name and argument tokens one by one (Figure~\ref{fig:task}). 
We evaluate seven models ranging from 0.96M to 88.43M parameters, indexed from 1 to 7 by increasing parameter count. 
Model configurations, including the numbers of encoder and decoder layers, hidden dimensions, and attention heads, are detailed in Appendix~\ref{appendix:model}.

\textbf{Dataset }
All experiments in this paper use $6\times6$ boards with cell values independently sampled from $\{0,\ldots,9\}$.
For clarity, however, we use $4\times4$ boards in all figures. 
Each instance contains an initial board, a target board, and a reference sequence that transforms the former into the latter.
We construct a TranSGrid evaluation set of 4,800 instances, marginally balanced across reference sequence lengths $L\in[1,12]$ and numbers of induction rules $K\in[0,3]$. 
Based on $L$, we divide the set into Easy ($L\in[1,4]$), Medium ($L\in[5,8]$), and Hard ($L\in[9,12]$), each containing 1,600 instances. 
Appendix~\ref{appendix:dataset} provides details on dataset construction and the joint $L$--$K$ distribution.

\textbf{Implementation details }
The two boards and their spatial relations can be encoded in multiple ways. 
We therefore evaluate four combinations of board and positional encodings and use the best overall configuration, \texttt{GRID}+\texttt{PAIR}, in all experiments (see Appendix~\ref{appendix:enc} for a detailed comparison).
All seven models are trained for eight epochs on 30M instances with reference sequence lengths ranging from 1 to 9.
For each model, we select the checkpoint with the highest solve rate on a 2,700 instance development set. 
Appendix~\ref{appendix:train_gen} details the generation procedure for the training and development sets. 
Using the same procedure, we construct a 4,800 instance Test set. 
Its reference sequence lengths are uniformly distributed from 1 to 12, and no Test instance appears in either the training or development set. 
We report results under greedy decoding (Greedy) and beam search (Top@8); for the latter, an instance is considered solved if any of the eight candidates is correct.

\subsection{Main Results}
\label{subsec:tg_result}

\begin{table}[ht]
\caption{Solve rates (\%) on the standard Test set and TranSGrid. 
The overall TranSGrid score is the unweighted mean over the Easy, Medium, and Hard subsets. 
Top@8 counts an instance as solved if any of its eight beam-search candidates succeeds upon execution.}
\centering
\small
\setlength{\tabcolsep}{2.8pt}
\makebox[\textwidth][c]{%
\begin{tabular}{@{}lccccccccccc@{}}
\toprule
& & \multicolumn{2}{c}{Test}
& \multicolumn{2}{c}{TranSGrid}
& \multicolumn{2}{c}{TranSGrid$_{\mathrm{Easy}}$}
& \multicolumn{2}{c}{TranSGrid$_{\mathrm{Medium}}$}
& \multicolumn{2}{c}{TranSGrid$_{\mathrm{Hard}}$} \\
\cmidrule(lr){3-4}
\cmidrule(lr){5-6}
\cmidrule(lr){7-8}
\cmidrule(lr){9-10}
\cmidrule(lr){11-12}
ID & Params (M)
& Greedy & Top@8
& Greedy & Top@8
& Greedy & Top@8
& Greedy & Top@8
& Greedy & Top@8 \\
\midrule
1 &  0.96 & 25.81 & 36.38 & 16.10 & 20.25 & 46.63 & 57.06 &  1.44 &  3.06 &  0.25 &  0.63 \\
2 &  1.89 & 35.79 & 51.54 & 18.10 & 23.77 & 48.94 & 61.56 &  4.88 &  8.63 &  0.50 &  1.13 \\
3 &  7.45 & 61.85 & 64.12 & 26.83 & 33.02 & 64.56 & 76.31 & 12.88 & 18.38 &  3.06 &  4.38 \\
4 & 11.13 & 71.27 & 72.50 & 32.06 & 40.85 & 71.13 & 86.69 & 19.69 & 28.19 &  5.38 &  7.69 \\
5 & 44.29 & 73.90 & 75.15 & 37.73 & 47.90 & 78.13 & 94.00 & 27.94 & 39.13 &  7.13 & 10.56 \\
6 & 59.00 & 78.33 & 78.90 & 50.40 & 58.56 & 92.06 & 97.94 & 45.00 & 59.75 & 14.13 & 18.00 \\
7 & 88.43 & 79.63 & 80.17 & 55.31 & 61.60 & 95.94 & 98.38 & 54.25 & 65.19 & 15.75 & 21.25 \\
\bottomrule
\end{tabular}}
\label{tab:tg_main}
\end{table}

As shown in Table~\ref{tab:tg_main}, models with IDs 4--7 achieve greedy solve rates of 71.27\%--79.63\% on Test, suggesting strong generalization under standard held-out evaluation. 
Across all seven models and both decoding strategies, however, TranSGrid solve rates are significantly lower than those on Test and decline sharply from Easy to Hard. 
Larger models and Top@8 improve solve rates, but performance remains limited on the more difficult subsets: under Top@8, even ID~7 achieves only 65.19\% on TranSGrid$_{\mathrm{Medium}}$ and 21.25\% on TranSGrid$_{\mathrm{Hard}}$.

\begin{figure}[ht]
\centering
\includegraphics[width=0.99\linewidth]{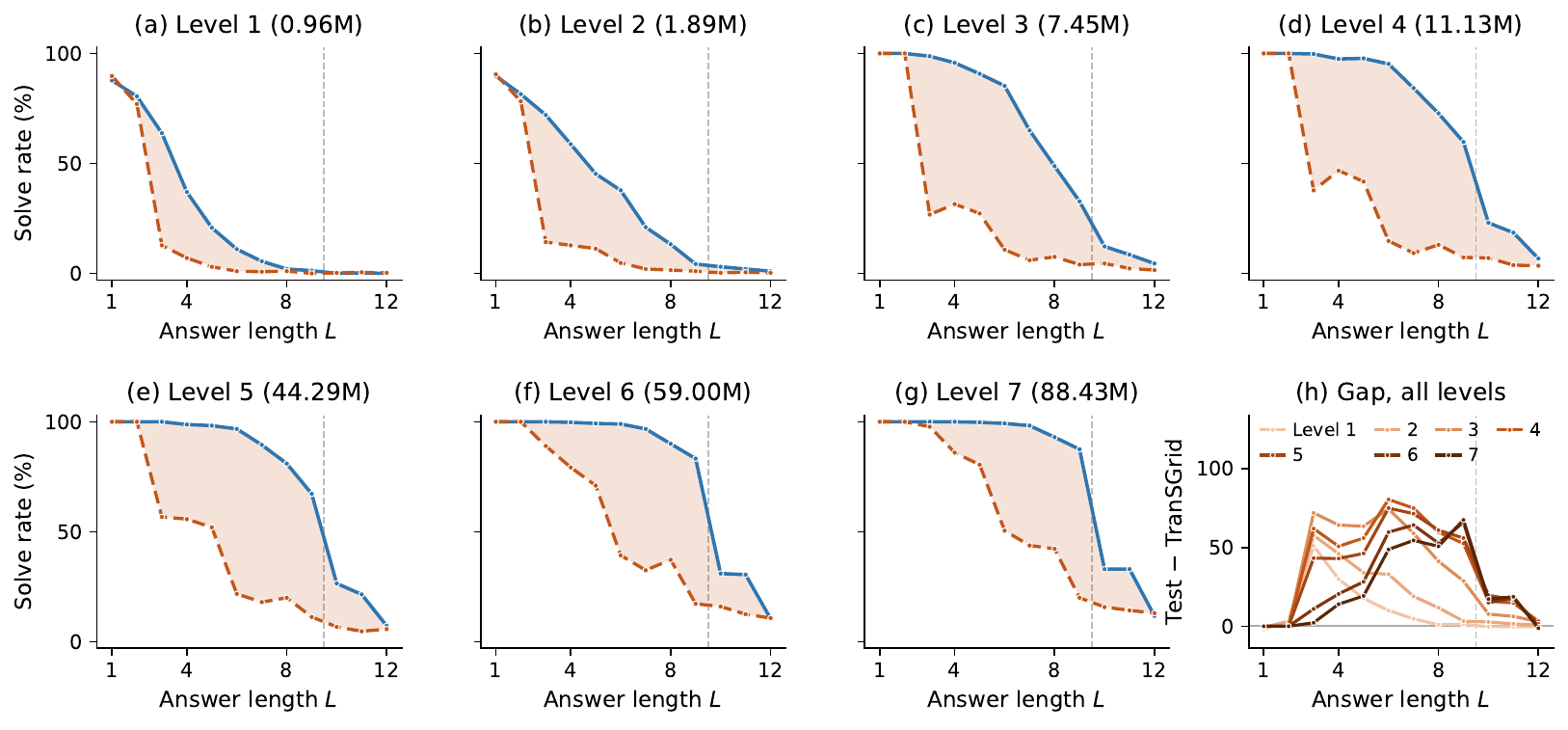}
\caption{Solve rate (\%) versus reference sequence length $L$ under greedy decoding.
Panels~(a)--(g) show results for model IDs 1--7, respectively.
In each panel, \swline{tgtest} and \swdash{tgrules} show performance on Test and TranSGrid, respectively, while \swband{tgrules!35}{tgband} marks the gap between them.
\swgrey marks the maximum training length, $L=9$, with larger values requiring length extrapolation.
Panel~(h) summarizes the Test--TranSGrid gaps for all seven models, with darker curves indicating larger models. }
\label{fig:transgrid_test_compare}
\end{figure}

To examine whether this gap persists at matched reference sequence lengths, we compare Test and TranSGrid solve rates in Figure~\ref{fig:transgrid_test_compare}. 
A substantial gap remains for $L\leq9$, even though this range is fully covered during training and requires no length extrapolation. 
Across most of this range, larger models maintain high solve rates on Test, whereas their TranSGrid performance declines sharply as $L$ increases. 
Panel~(h) further shows that increasing model capacity shifts the largest gap toward longer sequences but does not eliminate it within the training-length range. 
For $L>9$, the gap narrows as Test performance also declines under length extrapolation. 
Together, these results demonstrate the value of TranSGrid as a task for systematic generalization, while also highlighting its unique advantage of not relying on productivity.

\section{What Existing Metrics Capture and Miss:
A Reasoning- Centered Analysis}

\subsection{Inductive Reasoning Is Suppressed by Elemental Composition and Not Guaranteed by Productivity}
\label{subsec:reasoning_inductive}

Inductive reasoning involves discovering reusable composition rules from observed examples of how action effects interact and applying these rules to new instances. 
This demand is largely suppressed by elemental composition, where compound outcomes are explained mainly by individual action effects rather than interaction effects. 
Nor is it guaranteed by productivity: longer sequences may require only repeated applications of atomic actions, without requiring models to infer new composition rules. 
To test whether inductive reasoning contributes to systematic generalization beyond productivity and whether elemental composition suppresses this demand, we first analyze performance across inductive loads relative to $L$ and then compare TranSGrid with a decoupled variant. 

\begin{figure}[h]
\centering
\begin{subfigure}[b]{0.34\linewidth}
  \centering
  \includegraphics[width=\linewidth]{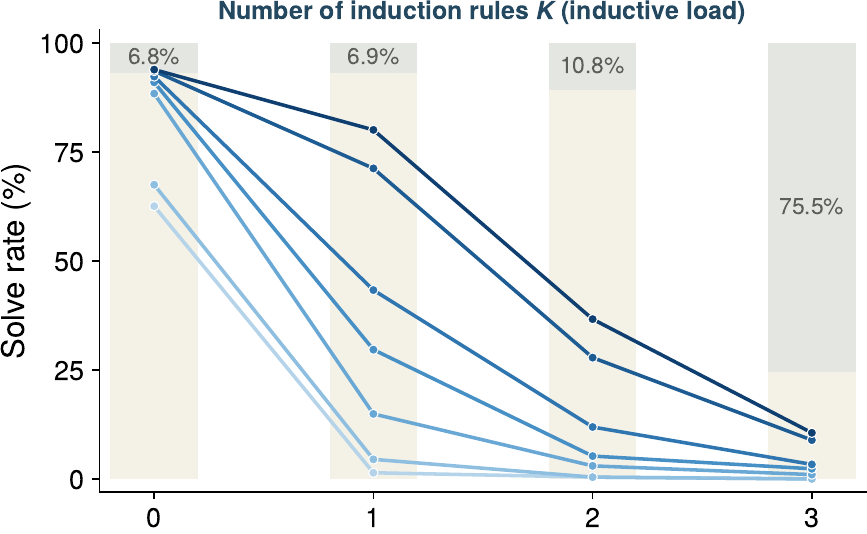}
  \vspace{-0.12\linewidth}
  \caption{}
  \label{fig:inductive}
\end{subfigure}\hfill%
\begin{subfigure}[b]{0.33\linewidth}
  \centering
  \includegraphics[width=\linewidth]{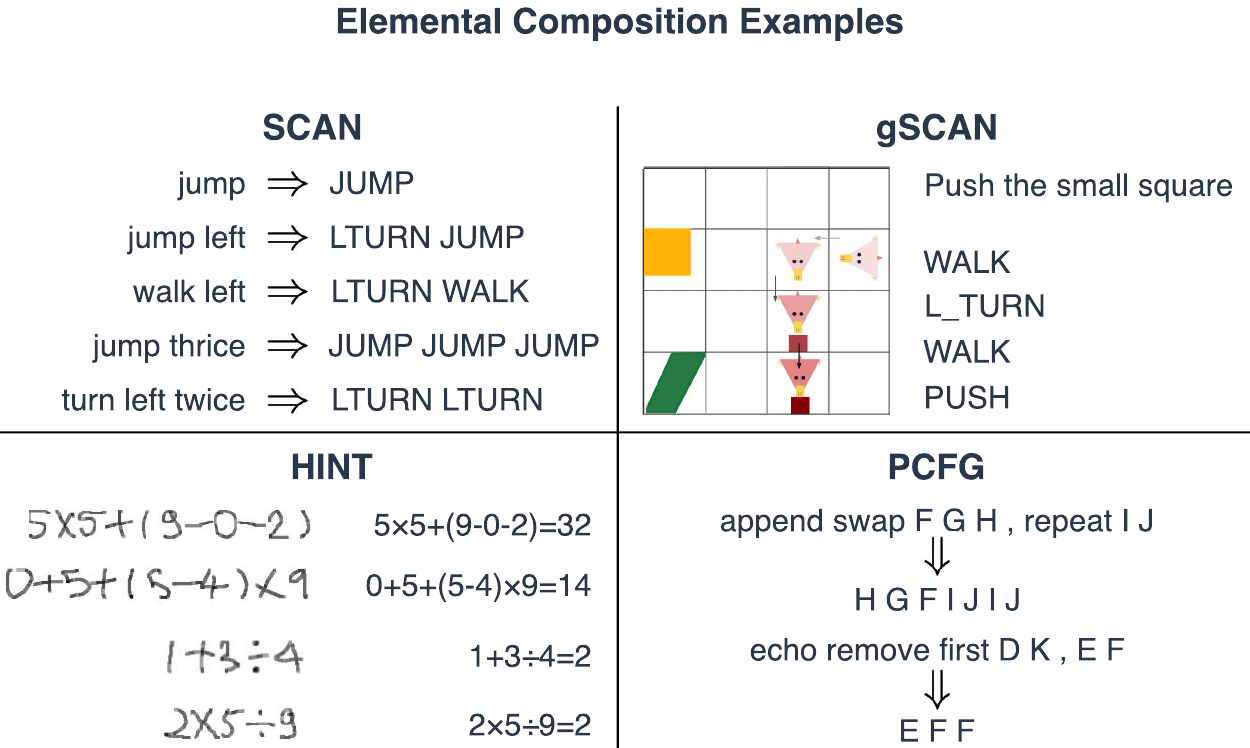}
  \caption{}
  \label{fig:lim_linear}
\end{subfigure}\hfill%
\begin{subfigure}[b]{0.27\linewidth}
  \centering
  \includegraphics[width=\linewidth]{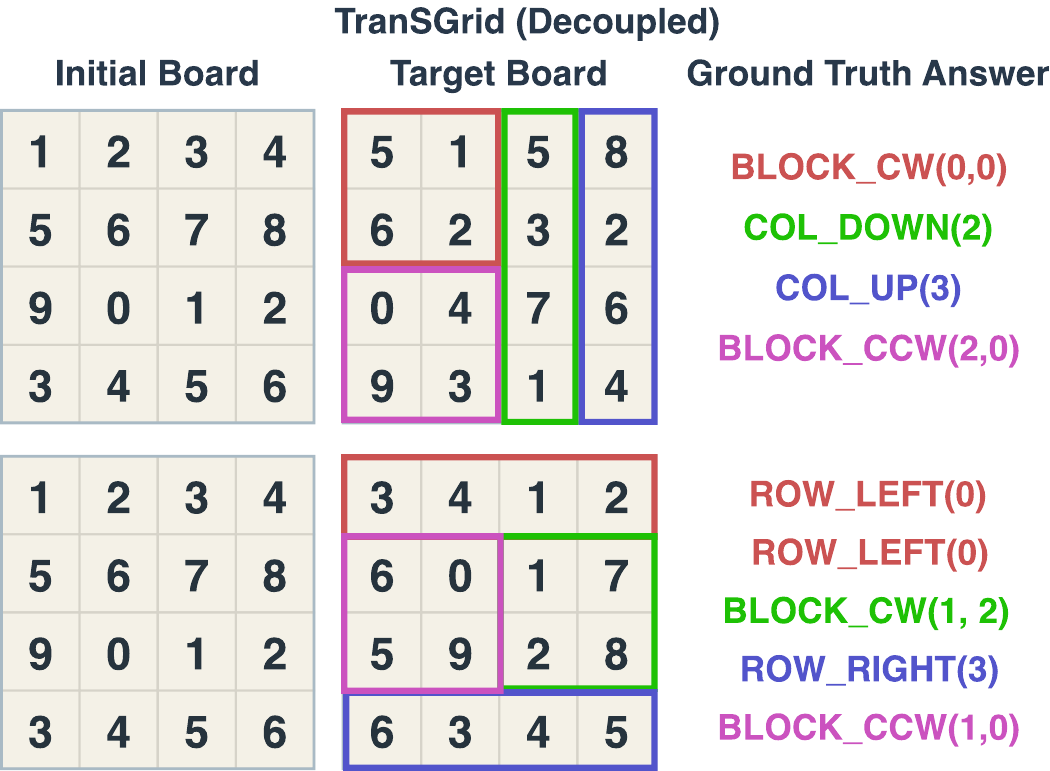}
  \caption{}
  \label{fig:transgrid_decoupled}
\end{subfigure}
\caption{(a) Solve rates on TranSGrid as inductive load $K$ increases. Curves from light to dark denote model IDs 1--7, and the shaded bars indicate the proportion of instances requiring length extrapolation.
(b) Examples of elemental composition in prior benchmarks, where interactions among atomic operations are limited.
(c) Examples from TranSGrid (Decoupled), where actions operate on separate parts of the board to reduce action interactions and approximate elemental composition.}

\label{fig:linear_inductive}
\end{figure}

Inductive reasoning presents a distinct challenge for systematic generalization that productivity alone does not capture. 
As shown in Figure~\ref{fig:inductive}, solve rates decline sharply across all seven models as inductive load \(K\) increases. 
More importantly, this decline is already pronounced even when productivity demands remain limited: as indicated by the shaded bars, only 10.8\% of the instances at $K=2$ require length extrapolation, yet solve rates are below 40\% for all models. 

\begin{table}[ht]
\caption{Comparison results on Test, TranSGrid, and TranSGrid (Decoupled). 
$\Delta$ denotes the absolute gain over the corresponding TranSGrid results.}
\label{tab:tg_linear}
\centering\small
\setlength{\tabcolsep}{5.2pt}
\makebox[\textwidth][c]{%
\begin{tabular}{@{}lccccccccc@{}}
\toprule
& & \multicolumn{2}{c}{Test} & \multicolumn{2}{c}{TranSGrid}
  & \multicolumn{2}{c}{TranSGrid (Decoupled)} & \multicolumn{2}{c}{$\Delta$} \\
\cmidrule(lr){3-4}\cmidrule(lr){5-6}\cmidrule(lr){7-8}\cmidrule(lr){9-10}
ID & Params (M) & Greedy & Top@8 & Greedy & Top@8 & Greedy & Top@8
   & Greedy & Top@8 \\
\midrule
1 &  0.96 & 25.81 & 36.38 & 16.10 & 20.25 & 23.23 & 46.77 & 7.13 & 26.52 \\
2 &  1.89 & 35.79 & 51.54 & 18.10 & 23.77 & 26.54 & 58.37 & 8.44 & 34.60 \\
3 &  7.45 & 61.85 & 64.12 & 26.83 & 33.02 & 70.79 & 73.06 & 43.96 & 40.04 \\
4 & 11.13 & 71.27 & 72.50 & 32.06 & 40.85 & 74.90 & 76.00 & 42.84 & 35.15 \\
5 & 44.29 & 73.90 & 75.15 & 37.73 & 47.90 & 75.90 & 76.98 & 38.17 & 29.08 \\
6 & 59.00 & 78.33 & 78.90 & 50.40 & 58.56 & 77.96 & 79.04 & 27.56 & 20.48 \\
7 & 88.43 & 79.63 & 80.17 & 55.31 & 61.60 & 78.42 & 79.87 & 23.11 & 18.27 \\
\bottomrule
\end{tabular}}
\end{table}

However, elemental composition largely suppresses this inductive demand by limiting interactions among action effects. 
Figure~\ref{fig:lim_linear} illustrates this structure in SCAN~\citep{lake2018generalization}, gSCAN~\citep{ruis2020benchmark}, HINT~\citep{li2023a}, and PCFG~\citep{hupkes2020compositionality}. 
To approximate elemental composition in a controlled setting, we construct TranSGrid (Decoupled), as shown in Figure~\ref{fig:transgrid_decoupled}. 
Specifically, TranSGrid (Decoupled) is constructed so that actions operate on separate parts of the board, thereby minimizing interference among their effects. 
We apply this decoupling to instances with $L\leq9$, which require no length extrapolation, while leaving instances with $L\geq10$ unchanged and preserving the initial boards and per-length counts. 

As shown in Table~\ref{tab:tg_linear}, decoupling consistently improves solve rates across all model sizes and both decoding strategies.
Gains reach 43.96 percentage points under greedy decoding and 40.04 points under Top@8. 
Performance recovers to roughly the Test level or higher for model IDs~3--7 under both decoding strategies.
For example, the largest model improves from 55.31\% to 78.42\% under greedy decoding and from 61.60\% to 79.87\% under Top@8, approaching its Test solve rates of 79.63\% and 80.17\%, respectively.
Because the reference action sequence length distribution remains unchanged, this recovery reflects reduced action interactions rather than reduced productivity demands.
Overall, these results show that elemental composition suppresses this inductive demand, and that productivity alone does not guarantee this form of reasoning.

\subsection{Action-Explicit Goals Largely Bypass Abductive Reasoning}
\label{subsec:reasoning_abductive}
Abductive reasoning in TranSGrid involves inferring, from the initial and target boards, which actions could have produced the observed transformation and in what order. 
Action-explicit goals largely suppress this demand by providing the complete solution sequence, including the actions and their order, in the input. 
Although the model still needs to generate the complete sequence, its task is largely reduced to mapping the input-specified composition to the corresponding output rather than inferring a suitable composition from scratch. 
To examine the role of abductive reasoning in systematic generalization and the effect of action-explicit goals, we first analyze how model performance varies with abductive load and then compare TranSGrid with an action-explicit variant.

\begin{figure}[h]
\centering
\begin{subfigure}[b]{0.34\linewidth}
  \centering
  \includegraphics[width=\linewidth]{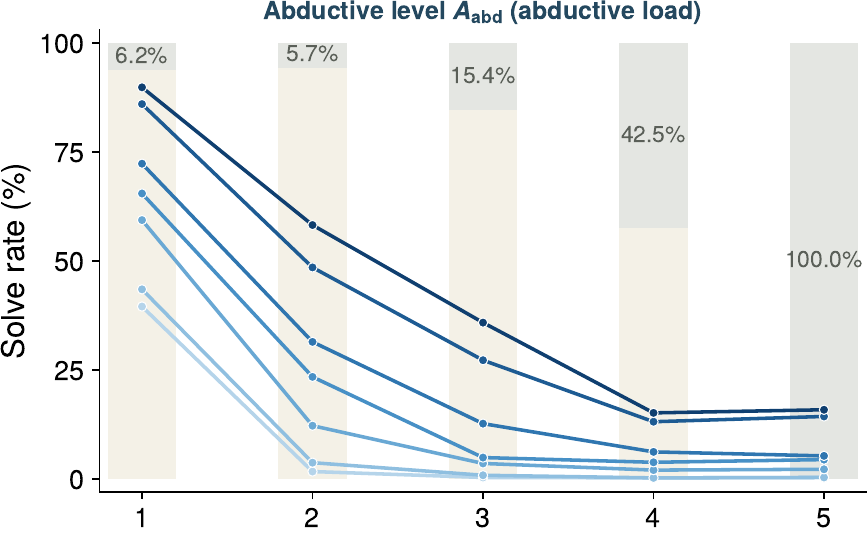}
  \vspace{-0.11\linewidth}
  \caption{}
  \label{fig:abductive_load}
\end{subfigure}\hfill%
\begin{subfigure}[b]{0.36\linewidth}
  \centering
  \includegraphics[width=\linewidth]{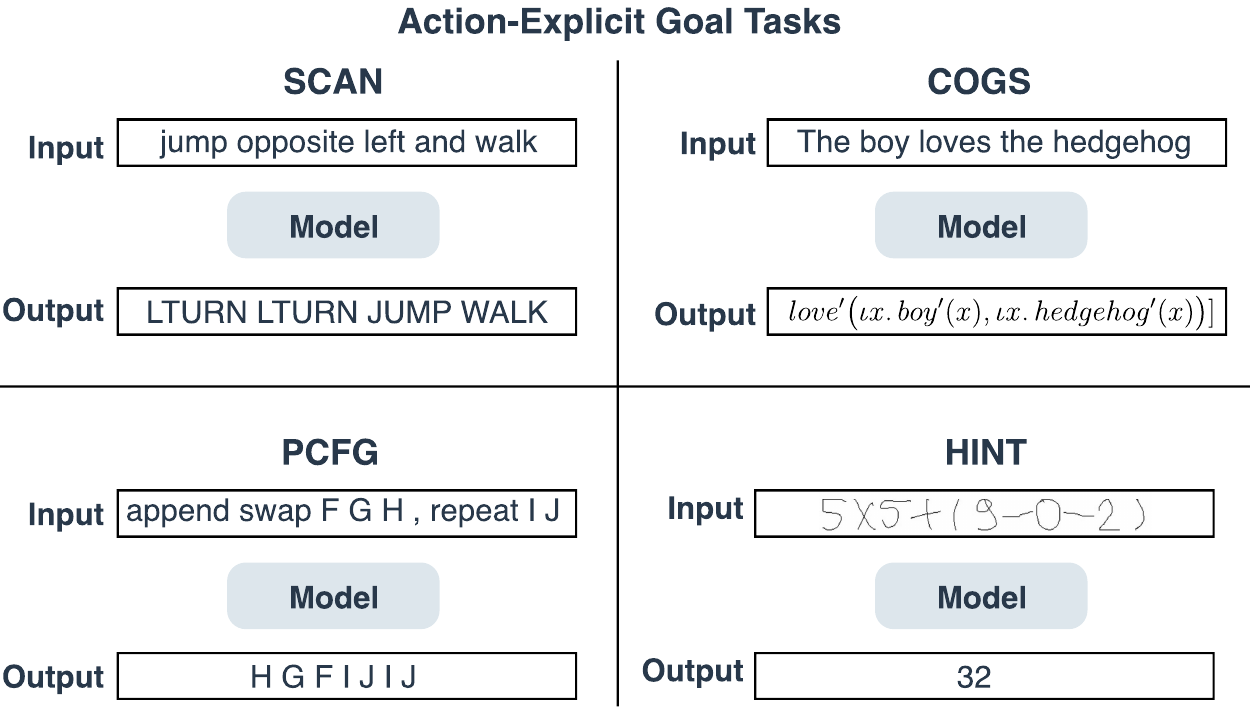}
  \caption{}
  \label{fig:abductive_task}
\end{subfigure}\hfill%
\begin{subfigure}[b]{0.245\linewidth}
  \centering
  \includegraphics[width=\linewidth]{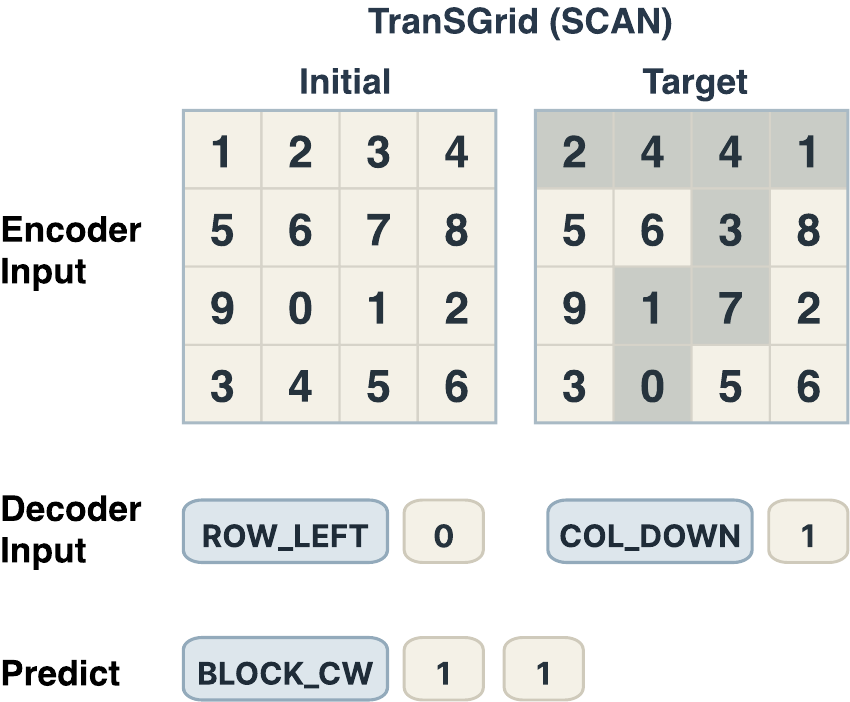}
  \caption{}
  \label{fig:transgrid_scan}
\end{subfigure}
\caption{
(a) Solve rates on TranSGrid as abductive load $A_{\mathrm{abd}}$ increases. Curves from light to dark denote model IDs 1--7, and the shaded bars indicate the proportion of instances requiring length extrapolation.
(b) Examples of action-explicit tasks in prior work, where the input specifies the complete action sequence to parse or sovle.
(c) TranSGrid (SCAN), which likewise requires the model to parse or execute an almost complete action sequence before producing the final answer.}
\label{fig:abductive}
\end{figure} 

As presented in Figure~\ref{fig:abductive_load}, solve rates decline sharply across all seven models as abductive load increases. 
Notably, between abductive levels 1 and 2, solve rates fall substantially even as the proportion of instances requiring length extrapolation decreases from 6.2\% to 5.7\%. 
These results indicate that abductive reasoning is an important dimension of systematic generalization that productivity alone does not fully capture.

Despite its importance, many prior benchmarks, such as SCAN~\citep{lake2018generalization}, COGS~\citep{kim2020cogs}, PCFG~\citep{hupkes2020compositionality}, and HINT~\citep{li2023a}, evaluate systematic generalization through action-explicit tasks, in which the input specifies the complete action sequence (Figure~\ref{fig:abductive_task}). 
Such tasks directly evaluate whether a model can parse or execute a novel composition, but have the side effect of reducing the need for abductive reasoning in systematic generalization. 
To examine the effect of this design in a controlled setting, we construct an action-explicit variant of TranSGrid, denoted TranSGrid (SCAN), as shown in Figure~\ref{fig:transgrid_scan}. 
TranSGrid (SCAN) leaves the encoder input unchanged but provides the decoder with the first $L-1$ gold actions, requiring the model to parse or execute an almost complete action sequence before producing the final answer, thereby substantially reducing the abductive demand. 

\begin{table}[ht]
\caption{Comparison results on Test, TranSGrid, and TranSGrid (SCAN). 
$\Delta$ denotes the absolute gain over the corresponding TranSGrid result.}
\label{tab:tg_scan}
\centering\small
\setlength{\tabcolsep}{6.1pt}
\makebox[\textwidth][c]{%
\begin{tabular}{@{}lccccccccc@{}}
\toprule
& & \multicolumn{2}{c}{Test} & \multicolumn{2}{c}{TranSGrid}
  & \multicolumn{2}{c}{TranSGrid (SCAN)} & \multicolumn{2}{c}{$\Delta$} \\
\cmidrule(lr){3-4}\cmidrule(lr){5-6}\cmidrule(lr){7-8}\cmidrule(lr){9-10}
ID & Params (M) & Greedy & Top@8 & Greedy & Top@8 & Greedy & Top@8
   & Greedy & Top@8 \\
\midrule
1 &  0.96 & 25.81 & 36.38 & 16.10 & 20.25 & 32.69 & 62.00 & 16.59 & 41.75 \\
2 &  1.89 & 35.79 & 51.54 & 18.10 & 23.77 & 37.90 & 69.65 & 19.80 & 45.88 \\
3 &  7.45 & 61.85 & 64.12 & 26.83 & 33.02 & 45.21 & 73.62 & 18.38 & 40.60 \\
4 & 11.13 & 71.27 & 72.50 & 32.06 & 40.85 & 53.02 & 74.73 & 20.96 & 33.88 \\
5 & 44.29 & 73.90 & 75.15 & 37.73 & 47.90 & 60.79 & 72.81 & 23.06 & 24.91 \\
6 & 59.00 & 78.33 & 78.90 & 50.40 & 58.56 & 62.31 & 72.85 & 11.91 & 14.29 \\
7 & 88.43 & 79.63 & 80.17 & 55.31 & 61.60 & 65.29 & 78.35 & 9.98 & 16.75 \\
\bottomrule
\end{tabular}}
\end{table}

Table~\ref{tab:tg_scan} shows that providing the gold action prefix consistently improves solve rates across all model sizes and both decoding strategies. 
The gains reach 23.06 percentage points for model ID~5 under greedy decoding and 45.88 points for model ID~2 under Top@8. 
Under Top@8, the smallest model improves from 20.25\% to 62.00\%, surpassing its Test result of 36.38\%, while the largest model improves from 61.60\% to 78.35\%, approaching its Test result of 80.17\%. 
This recovery arises primarily from reduced abductive demand rather than from the shorter output length: although the model only needs to predict the final action, it still needs to execute the given prefix and track intermediate states before producing the answer, as in prior action-explicit tasks. 
Taken together, our results show that tasks with action-explicit goals largely reduce the need for abductive reasoning and thus provide an incomplete picture of systematic generalization.

\subsection{A Reasoning-Centered Lens on Systematic Generalization}
\label{subsec:reasoning_relation}

The preceding analyses reveal two important blind spots in existing work on systematic generalization. 
Elemental composition suppresses the inductive demand of generalizing from observed action interactions to reusable composition patterns applicable to new task instances, whereas action-explicit goals bypass the abductive demand of inferring a solution sequence from the desired outcome.
Many prior studies rely on productivity as primary evidence of systematic generalization. 
Our results show, however, that productivity is neither necessary for systematic generalization nor sufficient to capture its full reasoning demands.

A comprehensive evaluation should therefore go beyond productivity and engage all three forms of reasoning. 
Specifically, deduction derives consequences from given rules and states; induction discovers reusable rules from observed examples; and abduction infers plausible action sequences that can achieve the desired outcome. 
Despite their distinct roles, these reasoning forms are closely coupled: each composition rule involves multiple actions, while longer sequences provide more opportunities for action interactions and concealment. 
This interplay is reflected in TranSGrid: longer sequences increase deductive load and provide more opportunities to introduce inductive and abductive demands, while reducing either of the latter makes the task much easier. 
We hope that, by capturing these coupled reasoning demands, TranSGrid could advance research on the mechanisms of systematic generalization and support the development of algorithms to improve this capability.

\section{Conclusion} 

This paper examines what common simplifications leave untested in systematic generalization through a reasoning-centered lens. 
To support this analysis, we introduce TranSGrid, a controllable testbed that jointly engages deduction, induction, and abduction within a unified task. 
Experiments with seven Transformer models on TranSGrid show that solve rates are much lower than those on the standard held-out test set. 
This gap persists on instances whose reference sequence lengths are within the training range, indicating that productivity alone is insufficient to evaluate systematic generalization. 
We further introduce TranSGrid (Decoupled) to approximate elemental composition and TranSGrid (SCAN) to make goals largely action-explicit. 
Both variants improve performance to roughly the held-out test level, suggesting that these simplifications reduce the inductive and abductive demands of systematic generalization. 
Therefore, a comprehensive evaluation of systematic generalization should go beyond productivity and jointly engage all three forms of reasoning.

\subsection*{AI use statement}

In this work, we used generative AI tools for designing or providing feedback on research methodology or experiments, implementing methods, and assisting with translation. 
We have not used generative AI tools for generating synthetic datasets, helping develop theoretical models or conceptual frameworks, formulating mathematical claims, proposing or refining hypotheses, cleaning and reformatting datasets, supporting qualitative and thematic data analysis, or interpreting results, and providing critical ingredients for proving mathematical claims and assisting in the writing of proofs are not applicable to this work. 
Additionally, we used generative AI tools for creating or editing software code, drafting parts of a research paper, summarizing or analyzing existing literature, sourcing/searching for information, editing the paper to improve readability, and proposing a title or keywords for the paper. 
We have reviewed all AI-assisted work: LLM-generated code was checked and verified for correctness by the authors; all reported results come from experiments we ran and inspected ourselves; AI-polished text was manually reviewed for factual accuracy, originality, and consistency with our intended claims. 
We take responsibility for the final content of this work, including text, claims or artifacts produced with the aid of generative AI.


\subsection*{Ethics statement}

All datasets used in this work are generated programmatically from numerical grids and predefined transformation rules. 
The study does not involve human participants or the collection or use of personal or sensitive data. 
Our experiments are conducted entirely within an abstract grid-transformation environment and aim to advance the understanding of systematic generalization. 
Given the scope of the data, task, and experiments, we do not identify any specific ethical concerns arising from this work. 
We plan to release the datasets and code for data generation, training, and evaluation to support reproducibility and further research.  

%

\subsection*{Reproducibility statement}

For our submission, we have uploaded the entirety of the source code as a zipped file that has been properly anonymized. 
The source code contains inline documentation that details purpose and usage of different parts of the codebase. 
In addition, we also include the full set of model responses and the evaluation script. 
As discussed in the ethics statement, we plan to more formally release TranSGrid to the public as an open source repository with thorough details that describes the framework, outlines the code, and details its usage.

%

%

\bibliography{iclr2027_conference}

@inproceedings{lake2018generalization,
  title={Generalization without systematicity: On the compositional skills of sequence-to-sequence recurrent networks},
  author={Lake, Brenden and Baroni, Marco},
  booktitle={International conference on machine learning},
  pages={2873--2882},
  year={2018},
  organization={PMLR}
}

@article{lake2023human,
  title={Human-like systematic generalization through a meta-learning neural network},
  author={Lake, Brenden M and Baroni, Marco},
  journal={Nature},
  volume={623},
  number={7985},
  pages={115--121},
  year={2023},
  publisher={Nature Publishing Group UK London}
}

@article{ruis2020benchmark,
  title={A benchmark for systematic generalization in grounded language understanding},
  author={Ruis, Laura and Andreas, Jacob and Baroni, Marco and Bouchacourt, Diane and Lake, Brenden M},
  journal={Advances in neural information processing systems},
  volume={33},
  pages={19861--19872},
  year={2020}
}

@inproceedings{wu2021reascan,
title={Rea{SCAN}: Compositional Reasoning in Language Grounding},
author={Zhengxuan Wu and Elisa Kreiss and Desmond Ong and Christopher Potts},
booktitle={Thirty-fifth Conference on Neural Information Processing Systems Datasets and Benchmarks Track (Round 1)},
year={2021},
url={https://openreview.net/forum?id=Rtquf4Jk0jN}
}

@inproceedings{sikarwar2022can,
  title={When can transformers ground and compose: Insights from compositional generalization benchmarks},
  author={Sikarwar, Ankur and Patel, Arkil and Goyal, Navin},
  booktitle={Proceedings of the 2022 Conference on Empirical Methods in Natural Language Processing},
  pages={648--669},
  year={2022}
}

@article{hupkes2020compositionality,
  title={Compositionality decomposed: How do neural networks generalise?},
  author={Hupkes, Dieuwke and Dankers, Verna and Mul, Mathijs and Bruni, Elia},
  journal={Journal of Artificial Intelligence Research},
  volume={67},
  pages={757--795},
  year={2020}
}

@inproceedings{li2023a,
title={A Minimalist Dataset for Systematic Generalization of Perception, Syntax, and Semantics},
author={Qing Li and Siyuan Huang and Yining Hong and Yixin Zhu and Ying Nian Wu and Song-Chun Zhu},
booktitle={The Eleventh International Conference on Learning Representations },
year={2023},
url={https://openreview.net/forum?id=kIPyTuEZuAK}
}

@inproceedings{kim2020cogs,
  title={COGS: A compositional generalization challenge based on semantic interpretation},
  author={Kim, Najoung and Linzen, Tal},
  booktitle={Proceedings of the 2020 conference on empirical methods in natural language processing (emnlp)},
  pages={9087--9105},
  year={2020}
}

@article{cloos2024baba,
  title={Baba is ai: Break the rules to beat the benchmark},
  author={Cloos, Nathan and Jens, Meagan and Naim, Michelangelo and Kuo, Yen-Ling and Cases, Ignacio and Barbu, Andrei and Cueva, Christopher J},
  journal={arXiv preprint arXiv:2407.13729},
  year={2024}
}

@inproceedings{kumon2025analyzing,
  title={Analyzing the inner workings of transformers in compositional generalization},
  author={Kumon, Ryoma and Yanaka, Hitomi},
  booktitle={Proceedings of the 2025 Conference of the Nations of the Americas Chapter of the Association for Computational Linguistics: Human Language Technologies (Volume 1: Long Papers)},
  pages={8529--8540},
  year={2025}
}

@article{chen2026sage,
  title={SAGE-Eval: Evaluating LLMs for systematic generalizations of safety facts},
  author={Chen, Yueh-Han and Davidson, Guy and Lake, Brenden},
  journal={Advances in Neural Information Processing Systems},
  volume={38},
  year={2026}
}

@article{wu2023recogs,
  title={Recogs: How incidental details of a logical form overshadow an evaluation of semantic interpretation},
  author={Wu, Zhengxuan and Manning, Christopher D and Potts, Christopher},
  journal={Transactions of the Association for Computational Linguistics},
  volume={11},
  pages={1719--1733},
  year={2023},
  publisher={MIT Press One Broadway, 12th Floor, Cambridge, Massachusetts 02142, USA~…}
}

@inproceedings{jabbar-etal-2025-distinguishing,
    title = "Distinguishing fair from unfair compositional generalization tasks",
    author = "Jabbar, Ahmad  and
      Condoravdi, Cleo  and
      Potts, Christopher",
    editor = "Christodoulopoulos, Christos  and
      Chakraborty, Tanmoy  and
      Rose, Carolyn  and
      Peng, Violet",
    booktitle = "Findings of the Association for Computational Linguistics: EMNLP 2025",
    month = nov,
    year = "2025",
    address = "Suzhou, China",
    publisher = "Association for Computational Linguistics",
    url = "https://aclanthology.org/2025.findings-emnlp.1133/",
    doi = "10.18653/v1/2025.findings-emnlp.1133",
    pages = "20796--20807",
    ISBN = "979-8-89176-335-7"
}

@inproceedings{ji2023drugood,
  title={Drugood: Out-of-distribution dataset curator and benchmark for ai-aided drug discovery--a focus on affinity prediction problems with noise annotations},
  author={Ji, Yuanfeng and Zhang, Lu and Wu, Jiaxiang and Wu, Bingzhe and Li, Lanqing and Huang, Long-Kai and Xu, Tingyang and Rong, Yu and Ren, Jie and Xue, Ding and others},
  booktitle={Proceedings of the AAAI Conference on Artificial Intelligence},
  volume={37},
  pages={8023--8031},
  year={2023}
}

@article{chen2026rule,
  title={Rule-Compliant Visual Spatial Planning for Multimodal Large Language Models},
  author={Chen, Yu and Lei, Ting and Li, Yaoyi and Cai, Jia and Wu, Zhecen and Liu, Yang},
  journal={arXiv preprint arXiv:2608.20237},
  year={2026}
}

@inproceedings{spilsbury-etal-2024-generating,
    title = "Generating Demonstrations for In-Context Compositional Generalization in Grounded Language Learning",
    author = "Spilsbury, Sam  and
      Marttinen, Pekka  and
      Ilin, Alexander",
    editor = "Al-Onaizan, Yaser  and
      Bansal, Mohit  and
      Chen, Yun-Nung",
    booktitle = "Proceedings of the 2024 Conference on Empirical Methods in Natural Language Processing",
    month = nov,
    year = "2024",
    address = "Miami, Florida, USA",
    publisher = "Association for Computational Linguistics",
    url = "https://aclanthology.org/2024.emnlp-main.893/",
    doi = "10.18653/v1/2024.emnlp-main.893",
    pages = "15960--15991"
}

@inproceedings{li2026speak,
  title={Speak-to-structure: Evaluating llms in open-domain natural language-driven molecule generation},
  author={Li, Jiatong and Li, Junxian and Wang, Weida and Liu, Yunqing and Zheng, Changmeng and Bian, Yatao and Zhou, Dongzhan and Wei, Xiao-Yong and Li, Qing},
  booktitle={Proceedings of the 32nd ACM SIGKDD Conference on Knowledge Discovery and Data Mining V. 2},
  pages={9314--9325},
  year={2026}
}

@article{li2026llms,
  title={Do LLMs Truly Generalize in the Molecular Domain? A Perturbation-Based Analysis},
  author={Li, Jiatong and Wang, Weida and Zheng, Changmeng and Zhang, Shufei and Bian, Yatao and Wei, Xiao-yong and Li, Qing},
  journal={arXiv preprint arXiv:2607.01800},
  year={2026}
}

@inproceedings{chen2025hierarchical,
title={Hierarchical Graph Tokenization for Molecule-Language Alignment},
author={Yongqiang Chen and Quanming Yao and Juzheng Zhang and James Cheng and Yatao Bian},
booktitle={Forty-second International Conference on Machine Learning},
year={2025},
url={https://openreview.net/forum?id=wpbNczwAwV}
}

@inproceedings{han2025progressive,
  title={Progressive compositionality in text-to-image generative models},
  author={Han, Xu and Jin, Linghao and Liu, Xiaofeng and Liang, Paul Pu},
  booktitle={International Conference on Learning Representations},
  volume={2025},
  pages={83268--83290},
  year={2025}
}

@article{huang2025t2i,
  title={T2i-compbench++: An enhanced and comprehensive benchmark for compositional text-to-image generation},
  author={Huang, Kaiyi and Duan, Chengqi and Sun, Kaiyue and Xie, Enze and Li, Zhenguo and Liu, Xihui},
  journal={IEEE Transactions on Pattern Analysis and Machine Intelligence},
  volume={47},
  number={5},
  pages={3563--3579},
  year={2025},
  publisher={IEEE}
}

@inproceedings{dat2025vsc,
  title={Vsc: Visual search compositional text-to-image diffusion model},
  author={Dat, Do Huu and Hyeon-Woo, Nam and Mao, Po-Yuan and Oh, Tae-Hyun},
  booktitle={Proceedings of the IEEE/CVF International Conference on Computer Vision},
  pages={19153--19162},
  year={2025}
}

@inproceedings{addepalli2025does,
  title={Does safety training of llms generalize to semantically related natural prompts?},
  author={Addepalli, Sravanti and Varun, Yerram and Suggala, Arun and Shanmugam, Karthikeyan and Jain, Prateek},
  booktitle={International Conference on Learning Representations},
  volume={2025},
  pages={43611--43631},
  year={2025}
}

@article{fu2026reinforcement,
  title={Reinforcement Learning for Compositional Generalization with Outcome-Level Optimization},
  author={Fu, Xiyan and Liu, Wei},
  journal={arXiv preprint arXiv:2605.04920},
  year={2026}
}

@inproceedings{mondorf2026compositionalarc,
title={Compositional-{ARC}: Assessing Systematic Generalization in Abstract Spatial Reasoning},
author={Philipp Mondorf and Shijia Zhou and Monica Riedler and Barbara Plank},
booktitle={The Fourteenth International Conference on Learning Representations},
year={2026},
url={https://openreview.net/forum?id=h497VpgFKd}
}

@book{peirce1934collected,
  title={Collected papers of charles sanders peirce},
  author={Peirce, Charles Sanders},
  volume={5},
  year={1934},
  publisher={Harvard University Press}
}

@article{shank1998extraordinary,
  title={The extraordinary ordinary powers of abductive reasoning},
  author={Shank, Gary},
  journal={Theory \& psychology},
  volume={8},
  number={6},
  pages={841--860},
  year={1998},
  publisher={Sage Publications Sage CA: Thousand Oaks, CA}
}

@article{chemero2026abduction,
  title={Abduction and deduction in dynamical cognitive science},
  author={Chemero, Anthony},
  journal={Topics in Cognitive Science},
  volume={18},
  number={3},
  pages={e12692},
  year={2026},
  publisher={Wiley Online Library}
}

@article{devaud2015neural,
  title={Neural substrate for higher-order learning in an insect: mushroom bodies are necessary for configural discriminations},
  author={Devaud, Jean-Marc and Papouin, Thomas and Carcaud, Julie and Sandoz, Jean-Christophe and Gr{\"u}newald, Bernd and Giurfa, Martin},
  journal={Proceedings of the National Academy of Sciences},
  volume={112},
  number={43},
  pages={E5854--E5862},
  year={2015},
  publisher={National Academy of Sciences}
}

@article{duncan2018more,
  title={More than the sum of its parts: a role for the hippocampus in configural reinforcement learning},
  author={Duncan, Katherine and Doll, Bradley B and Daw, Nathaniel D and Shohamy, Daphna},
  journal={Neuron},
  volume={98},
  number={3},
  pages={645--657},
  year={2018},
  publisher={Elsevier}
}

@article{rotshtein2007role,
  title={Role of features and second-order spatial relations in face discrimination, face recognition, and individual face skills: Behavioral and functional magnetic resonance imaging data},
  author={Rotshtein, Pia and Geng, Joy J and Driver, Jon and Dolan, Raymond J},
  journal={Journal of Cognitive Neuroscience},
  volume={19},
  number={9},
  pages={1435--1452},
  year={2007},
  publisher={MIT Press One Rogers Street, Cambridge, MA 02142-1209, USA journals-info~…}
}

@article{leong2023holistic,
  title={Holistic and featural processing’s link to face recognition varies by individual and task},
  author={Leong, Bryan Qi Zheng and Estudillo, Alejandro J and Hussain Ismail, Ahamed Miflah},
  journal={Scientific Reports},
  volume={13},
  number={1},
  pages={16869},
  year={2023},
  publisher={Nature Publishing Group UK London}
}
\bibliographystyle{iclr2027_conference}

\appendix

\section{Induction Rules}
\label{appendix:induction_rules}

An induction rule is a short program of three to five atomic actions whose effects cancel or overwrite one another, leaving a net transformation that permutes only a few cells within a small local window. 
Table~\ref{tab:induction-rules} lists the twelve induction rules used in this paper, together with their constituent action sequences and net effects. 
Here, $(r,c)$ denotes the top-left corner of the local window. 
For a block operation such as \texttt{BLOCK\_CW}$(r{+}i,c{+}j)$, the two arguments specify the top-left cell of the $2\times2$ block to be rotated, where $i$ and $j$ are offsets from the rule origin. 
For row and column operations, $r{+}i$ and $c{+}j$ specify the corresponding row and column indices, respectively.
Collectively, the twelve rules realize localized permutations of two to five cells, including pairwise swaps, three to five cell cycles, and two disjoint swaps arranged in horizontal, vertical, diagonal, L-shaped, or bent configurations. 
The actions in each program are executed from top to bottom. The final column illustrates the resulting net effect using a concrete example. 
For each example, the values of $r$ and $c$ are provided above the boards, and the cells involved in the permutation are highlighted in red.

\begingroup
\footnotesize
\setlength{\tabcolsep}{4pt}
\setlength{\LTcapwidth}{\textwidth}

\begin{longtable}{@{}lll@{}}
\caption{The twelve induction rules used in this paper, together with their constituent action sequences and illustrative net effects. 
In each illustration, the rule origin $(r,c)$ is indicated above the boards, and the cells permuted by the rule are highlighted in red.}
\label{tab:induction-rules}\\
\toprule
Induction Rule & \multicolumn{1}{c}{Action sequence} & \multicolumn{1}{c}{Illustrative net effect} \\
\midrule
\endfirsthead
\toprule
Induction Rule & \multicolumn{1}{c}{Action sequence} & \multicolumn{1}{c}{Illustrative net effect} \\
\midrule
\endhead
\multicolumn{3}{r@{}}{\emph{continued on the next page}}\\
\endfoot
\bottomrule
\endlastfoot

\parbox[c][3.1cm][c]{0.29\textwidth}{\raggedright\texttt{horizontal\_swap\_mixed}} &
\parbox[c][3.1cm][c]{0.27\textwidth}{\centering
  \shortstack[c]{
    \texttt{BLOCK\_CW}$(r,c)$\\
    \texttt{COL\_DOWN}$(c)$\\
    \texttt{ROW\_LEFT}$(r{+}1)$\\
    \texttt{COL\_UP}$(c)$\\
    \texttt{ROW\_RIGHT}$(r{+}1)$
  }
} &
\parbox[c][3.1cm][c]{0.36\textwidth}{\centering
  $(r,c)=(0,0)$\par\vspace{2pt}
\includegraphics[width=0.90\linewidth,height=2.35cm,keepaspectratio]{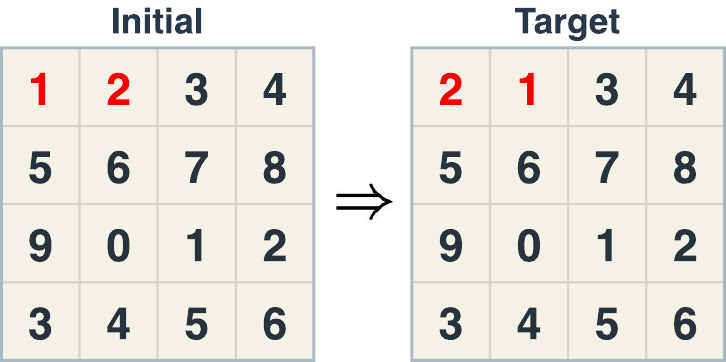}
} \\

\midrule

\parbox[c][3.1cm][c]{0.29\textwidth}{\raggedright\texttt{diagonal\_swap\_blocks}} &
\parbox[c][3.1cm][c]{0.27\textwidth}{\centering
  \shortstack[c]{
    \texttt{BLOCK\_CW}$(r,c{+}1)$\\
    \texttt{BLOCK\_CCW}$(r{+}1,c)$\\
    \texttt{BLOCK\_CCW}$(r,c)$\\
    \texttt{BLOCK\_CCW}$(r,c{+}1)$\\
    \texttt{BLOCK\_CW}$(r{+}1,c)$
  }
} &
\parbox[c][3.1cm][c]{0.36\textwidth}{\centering
  $(r,c)=(0,0)$\par\vspace{2pt}
\includegraphics[width=0.90\linewidth,height=2.35cm,keepaspectratio]{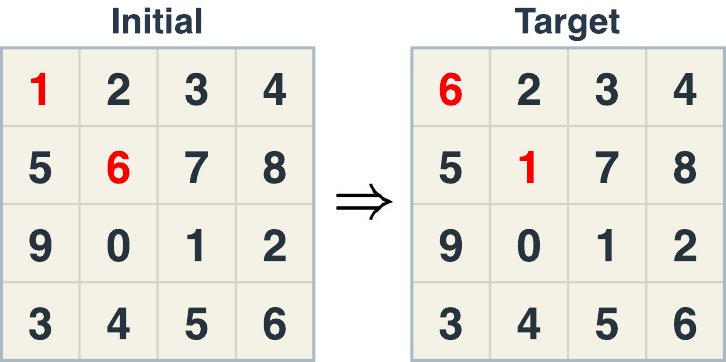}
} \\
\midrule

\parbox[c][3.1cm][c]{0.29\textwidth}{\raggedright\texttt{bent\_four\_cycle}} &
\parbox[c][3.1cm][c]{0.27\textwidth}{\centering
  \shortstack[c]{
    \texttt{BLOCK\_CW}$(r,c)$\\
    \texttt{BLOCK\_CW}$(r,c{+}1)$\\
    \texttt{BLOCK\_CCW}$(r,c)$
  }
} &
\parbox[c][3.1cm][c]{0.36\textwidth}{\centering
  $(r,c)=(0,0)$\par\vspace{2pt}
\includegraphics[width=0.90\linewidth,height=2.35cm,keepaspectratio]{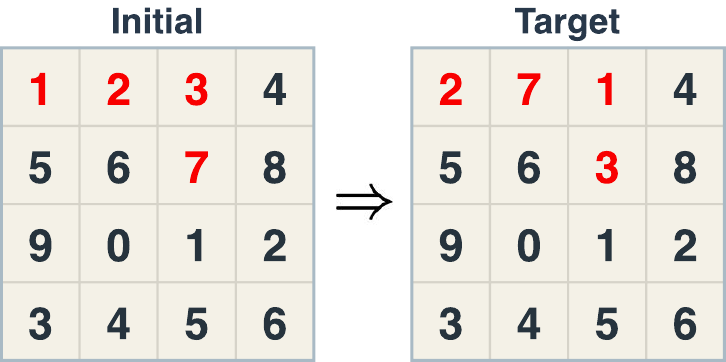}
} \\
\midrule

\parbox[c][3.1cm][c]{0.29\textwidth}{\raggedright\texttt{wide\_bent\_four\_cycle}} &
\parbox[c][3.1cm][c]{0.27\textwidth}{\centering
  \shortstack[c]{
    \texttt{BLOCK\_CW}$(r,c{+}1)$\\
    \texttt{BLOCK\_CW}$(r,c)$\\
    \texttt{BLOCK\_CCW}$(r,c{+}1)$
  }
} &
\parbox[c][3.1cm][c]{0.36\textwidth}{\centering
  $(r,c)=(0,0)$\par\vspace{2pt}
\includegraphics[width=0.90\linewidth,height=2.35cm,keepaspectratio]{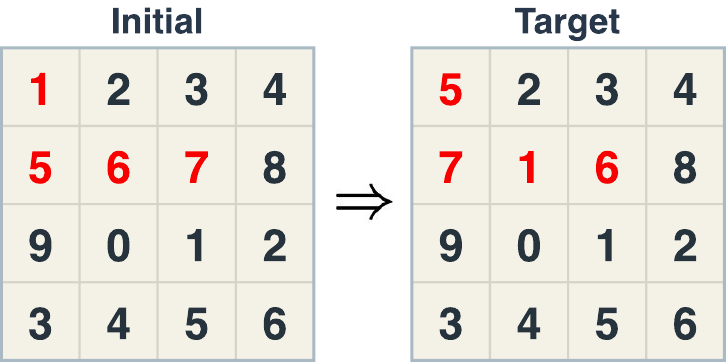}
} \\
\midrule

\parbox[c][3.1cm][c]{0.29\textwidth}{\raggedright\texttt{tall\_bent\_four\_cycle}} &
\parbox[c][3.1cm][c]{0.27\textwidth}{\centering
  \shortstack[c]{
    \texttt{BLOCK\_CW}$(r,c)$\\
    \texttt{BLOCK\_CW}$(r{+}1,c)$\\
    \texttt{BLOCK\_CCW}$(r,c)$
  }
} &
\parbox[c][3.1cm][c]{0.36\textwidth}{\centering
  $(r,c)=(0,0)$\par\vspace{2pt}
\includegraphics[width=0.90\linewidth,height=2.35cm,keepaspectratio]{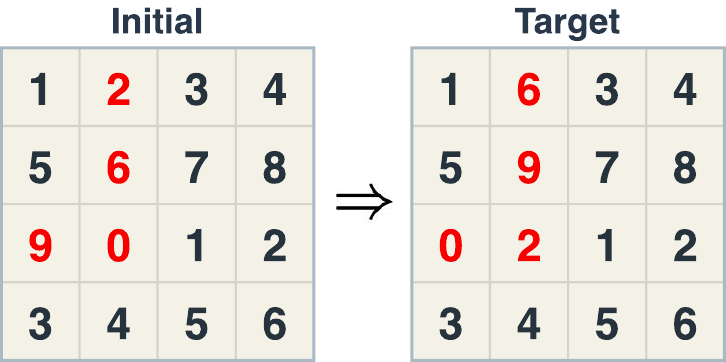}
} \\

\midrule

\parbox[c][3.1cm][c]{0.29\textwidth}{\raggedright\texttt{tall\_bent\_flipped}} &
\parbox[c][3.1cm][c]{0.27\textwidth}{\centering
  \shortstack[c]{
    \texttt{BLOCK\_CW}$(r{+}1,c)$\\
    \texttt{BLOCK\_CW}$(r,c)$\\
    \texttt{BLOCK\_CCW}$(r{+}1,c)$
  }
} &
\parbox[c][3.1cm][c]{0.36\textwidth}{\centering
  $(r,c)=(0,0)$\par\vspace{2pt}
\includegraphics[width=0.90\linewidth,height=2.35cm,keepaspectratio]{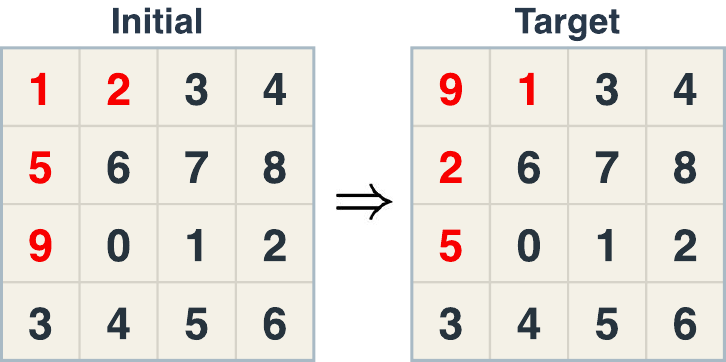}
} \\
\midrule

\parbox[c][3.1cm][c]{0.29\textwidth}{\raggedright\texttt{horizontal\_three\_cycle}} &
\parbox[c][3.1cm][c]{0.27\textwidth}{\centering
  \shortstack[c]{
    \texttt{ROW\_LEFT}$(r)$\\
    \texttt{COL\_LEFT}$(c{+}1)$\\
    \texttt{ROW\_RIGHT}$(r)$\\
    \texttt{COL\_LEFT}$(c{+}1)$
  }
} &
\parbox[c][3.1cm][c]{0.36\textwidth}{\centering
  $(r,c)=(0,0)$\par\vspace{2pt}
\includegraphics[width=0.90\linewidth,height=2.35cm,keepaspectratio]{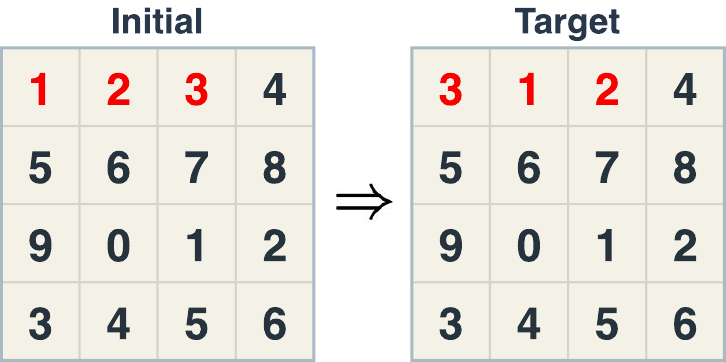}
} \\
\midrule

\parbox[c][3.1cm][c]{0.29\textwidth}{\raggedright\texttt{vertical\_three\_cycle}} &
\parbox[c][3.1cm][c]{0.27\textwidth}{\centering
  \shortstack[c]{
    \texttt{COL\_UP}$(c)$\\
    \texttt{ROW\_DOWN}$(r)$\\
    \texttt{COL\_DOWN}$(c)$\\
    \texttt{ROW\_DOWN}$(r)$
  }
} &
\parbox[c][3.1cm][c]{0.36\textwidth}{\centering
  $(r,c)=(0,0)$\par\vspace{2pt}
\includegraphics[width=0.90\linewidth,height=2.35cm,keepaspectratio]{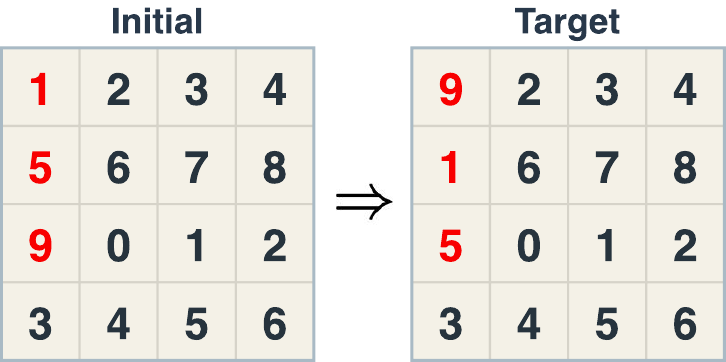}
} \\
\midrule

\parbox[c][3.1cm][c]{0.29\textwidth}{\raggedright\texttt{l\_three\_cycle}} &
\parbox[c][3.1cm][c]{0.27\textwidth}{\centering
  \shortstack[c]{
    \texttt{COL\_UP}$(c)$\\
    \texttt{ROW\_LEFT}$(r)$\\
    \texttt{COL\_DOWN}$(c)$\\
    \texttt{ROW\_RIGHT}$(r)$
  }
} &
\parbox[c][3.1cm][c]{0.36\textwidth}{\centering
  $(r,c)=(0,0)$\par\vspace{2pt}
\includegraphics[width=0.90\linewidth,height=2.35cm,keepaspectratio]{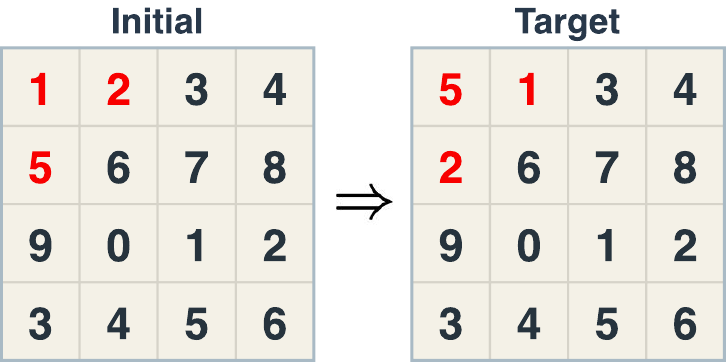}
} \\
\midrule

\parbox[c][3.1cm][c]{0.29\textwidth}{\raggedright\texttt{double\_horizontal\_swap}} &
\parbox[c][3.1cm][c]{0.27\textwidth}{\centering
  \shortstack[c]{
    \texttt{BLOCK\_CW}$(r,c)$\\
    \texttt{BLOCK\_CW}$(r,c{+}1)$\\
    \texttt{BLOCK\_CCW}$(r,c)$\\
    \texttt{BLOCK\_CCW}$(r,c{+}1)$
  }
} &
\parbox[c][3.1cm][c]{0.36\textwidth}{\centering
  $(r,c)=(0,0)$\par\vspace{2pt}
\includegraphics[width=0.90\linewidth,height=2.35cm,keepaspectratio]{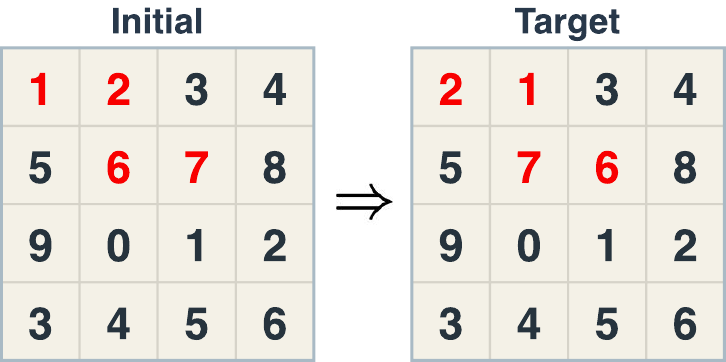}
} \\
\midrule

\parbox[c][3.1cm][c]{0.29\textwidth}{\raggedright\texttt{five\_cycle\_2x3}} &
\parbox[c][3.1cm][c]{0.27\textwidth}{\centering
  \shortstack[c]{
    \texttt{BLOCK\_CW}$(r,c)$\\
    \texttt{COL\_RIGHT}$(c{+}1)$\\
    \texttt{BLOCK\_CCW}$(r,c)$\\
    \texttt{COL\_RIGHT}$(c{+}1)$
  }
} &
\parbox[c][3.1cm][c]{0.36\textwidth}{\centering
  $(r,c)=(0,0)$\par\vspace{2pt}
\includegraphics[width=0.90\linewidth,height=2.35cm,keepaspectratio]{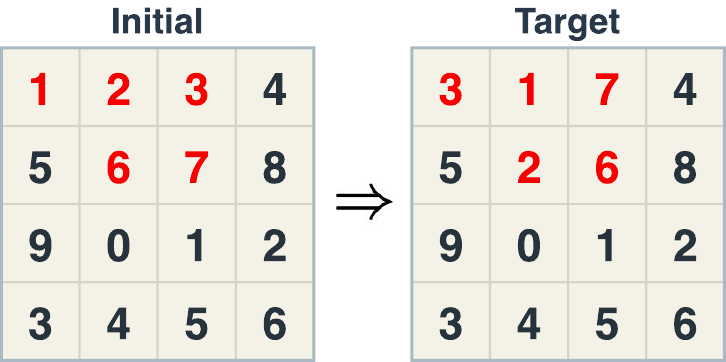}
} \\
\midrule

\parbox[c][3.1cm][c]{0.29\textwidth}{\raggedright\texttt{vertical\_swap\_blocks}} &
\parbox[c][3.1cm][c]{0.27\textwidth}{\centering
  \shortstack[c]{
    \texttt{BLOCK\_CW}$(r,c)$\\
    \texttt{BLOCK\_CW}$(r{+}1,c)$\\
    \texttt{BLOCK\_CCW}$(r,c{+}1)$\\
    \texttt{BLOCK\_CCW}$(r{+}1,c)$\\
    \texttt{BLOCK\_CW}$(r,c{+}1)$
  }
} &
\parbox[c][3.1cm][c]{0.36\textwidth}{\centering
  $(r,c)=(0,0)$\par\vspace{2pt}
\includegraphics[width=0.90\linewidth,height=2.35cm,keepaspectratio]{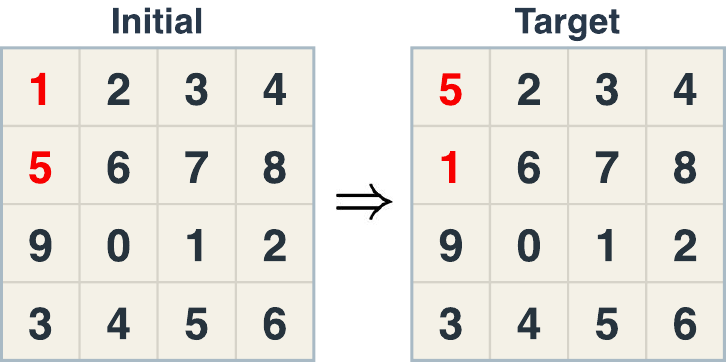}
} \\

\end{longtable}
\endgroup

\section{Estimating the Effective Action Length}
\label{appendix:effective_length}

Let $D$ denote the number of cells whose values differ between the initial and target boards. 
To convert $D$ into an effective action length, we first estimate the expected footprint of a random trajectory:
\[
g(\ell)=\mathbb{E}[D\mid \ell],
\]
where $\ell$ denotes the action-sequence length. 
For each $\ell\in\{1,\ldots,12\}$, we sample $400$ random trajectories of length $\ell$ on $6\times6$ boards whose cell values are drawn independently from $\{0,\ldots,9\}$. 
We execute each trajectory, count the cells whose values differ between its initial and final boards, and average these counts. 
The resulting footprint curve is reported in Table~\ref{tab:footprint}.

\begin{table}[ht]
\caption{The estimated footprint curve. Each value is averaged over 400 random trajectories of the corresponding length.}
\label{tab:footprint}
\centering
\small

\setlength{\tabcolsep}{6.5pt}

\makebox[\textwidth][c]{%
\begin{tabular}{@{}lcccccccccccc@{}}
\toprule
Length $\ell$
& 1 & 2 & 3 & 4 & 5 & 6
& 7 & 8 & 9 & 10 & 11 & 12 \\
\midrule
Footprint $g(\ell)$
& 7.2 & 12.5 & 16.0 & 19.1 & 21.3 & 22.9
& 24.4 & 25.0 & 26.3 & 27.0 & 27.3 & 27.8 \\
\bottomrule
\end{tabular}%
}
\end{table}

We obtain $L_{\mathrm{eff}}$ by inverting this curve through linear interpolation:
\[
L_{\mathrm{eff}}(D)=
\begin{cases}
\dfrac{D}{g(1)},
& 0\leq D\leq g(1), \\[6pt]
\ell+
\dfrac{D-g(\ell)}
      {g(\ell+1)-g(\ell)},
& g(\ell)<D\leq g(\ell+1),
  \quad \ell\in\{1,\ldots,11\}, \\[8pt]
12,
& D>g(12).
\end{cases}
\]
For $D\leq g(1)$, the effective length is interpolated between $(D,L_{\mathrm{eff}})=(0,0)$ and $(g(1),1)$. Values between adjacent entries of the footprint curve are linearly interpolated, while values above $g(12)$ are assigned an effective action length of $12$.

\section{Model Architectures}
\label{appendix:model}

We evaluate seven encoder-decoder Transformer models that vary in depth and hidden dimensions. 
In every configuration, the encoder and decoder contain the same number of layers. 
The encoder jointly processes the initial and target boards, while the decoder generates the action sequence autoregressively. 
Each action is represented by an action-name token followed by its argument tokens, and decoding terminates when the model generates the \texttt{EOS} token. 

\begin{table}[ht]
\caption{Details of the seven Transformer models used in this paper. 
Model IDs are ordered by the total number of trainable parameters.}
\label{tab:tg_model_ladder}
\centering
\small
\begin{tabular}{@{}lcccccr@{}}
\toprule
ID & Enc.\ layers & Dec.\ layers & $d_{\mathrm{model}}$
& Heads & $d_{\mathrm{ff}}$ & Params (M) \\
\midrule
1 & 2  & 2  & 128 & 2 & 512  & 0.96 \\
2 & 4  & 4  & 128 & 2 & 512  & 1.89 \\
3 & 4  & 4  & 256 & 4 & 1024 & 7.45 \\
4 & 6  & 6  & 256 & 4 & 1024 & 11.13 \\
5 & 6  & 6  & 512 & 8 & 2048 & 44.29 \\
6 & 8  & 8  & 512 & 8 & 2048 & 59.00 \\
7 & 12 & 12 & 512 & 8 & 2048 & 88.43 \\
\bottomrule
\end{tabular}
\end{table}

Table~\ref{tab:tg_model_ladder} summarizes the architectural configurations of the seven models. 
Here, $d_{\mathrm{model}}$ is the number of units in each Transformer bottleneck layer, $d_{\mathrm{ff}}=4d_{\mathrm{model}}$ is the number of units in each feed-forward sublayer, and Heads is the number of attention heads. 
The reported parameter counts include all trainable parameters under the \texttt{GRID}+\texttt{PAIR} encoding used in the main experiments.

\section{Dataset Construction and Composition}
\label{appendix:dataset}

\paragraph{Instance construction.}
Each instance begins with an initial board $B^{(0)}\in\mathcal{V}^{6\times6}$, whose cells are sampled independently from $\mathcal{V}=\{0,\ldots,9\}$.
We generate a reference action sequence $\pi=(a_1,\ldots,a_L)$ and execute it to obtain the target board:
\[
B^\star=T(\pi,B^{(0)}).
\]
The reference sequence guarantees the existence of a valid solution, but it need not be the shortest or unique solution. 
To construct an instance with inductive load $K$, we sample $K$ induction rules with replacement, subject to $\sum_{k=1}^{K}|\rho_k|\leq L$, and instantiate each rule at a valid board location.
We fill the remaining $L-\sum_{k=1}^{K}|\rho_k|$ positions with randomly sampled atomic actions.
The instantiated rules and individual filler actions are then shuffled as blocks, preserving the internal action order of each rule.
The induction rules are defined in Appendix~\ref{appendix:induction_rules}.

\paragraph{Dataset composition.}
The TranSGrid evaluation set contains 4,800 instances and is marginally balanced with respect to reference sequence length $L$ and inductive load $K$.
Each length $L\in\{1,\ldots,12\}$ contains 400 instances, and each load $K\in\{0,\ldots,3\}$ contains 1,200 instances.
Because every induction rule contains at least three actions, a feasible pair must satisfy $L\geq3K$.
Table~\ref{tab:tg_data_distribution} reports the resulting joint distribution.

\begin{table}[ht]
\caption{Joint distribution of the $4{,}800$ TranSGrid evaluation instances by reference sequence length $L$ and inductive load $K$.}
\label{tab:tg_data_distribution}
\centering
\small
\begin{tabular*}{0.88\textwidth}{@{\extracolsep{\fill}}llrrrrr@{}}
\toprule
\multirow{2}{*}{Subset}
& \multirow{2}{*}{Reference length $L$}
& \multicolumn{4}{c}{Inductive load $K$}
& \multirow{2}{*}{Total} \\
\cmidrule(lr){3-6}
& & $0$ & $1$ & $2$ & $3$ & \\
\midrule
\multirow{4}{*}{Easy}
& $1$ & 400 & --  & --  & --  & 400 \\
& $2$ & 400 & --  & --  & --  & 400 \\
& $3$ & 66  & 334 & --  & --  & 400 \\
& $4$ & 65  & 335 & --  & --  & 400 \\
\cmidrule{1-7}
\multirow{4}{*}{Medium}
& $5$ & 71 & 329 & --  & --  & 400 \\
& $6$ & 36 & 31  & 333 & --  & 400 \\
& $7$ & 16 & 26  & 358 & --  & 400 \\
& $8$ & 30 & 34  & 336 & --  & 400 \\
\cmidrule{1-7}
\multirow{4}{*}{Hard}
& $9$  & 35 & 28 & 43 & 294 & 400 \\
& $10$ & 34 & 30 & 39 & 297 & 400 \\
& $11$ & 24 & 27 & 46 & 303 & 400 \\
& $12$ & 23 & 26 & 45 & 306 & 400 \\
\midrule
\multicolumn{2}{@{}l}{Total}
& 1{,}200 & 1{,}200 & 1{,}200 & 1{,}200 & 4{,}800 \\
\bottomrule
\end{tabular*}
\end{table}

\section{Input Encoding Variants}
\label{appendix:enc}

\paragraph{Encoding variants.}
We compare two board encodings and two positional encoding schemes. 
Under the \texttt{PAIR} encoding, the values at corresponding locations of the initial and target boards are represented jointly by a single token, producing an input sequence of $36$ tokens. 
Under the \texttt{BOARD} encoding, the two boards are serialized separately, producing an input sequence of $86$ tokens. 
For positional encoding, \texttt{GRID} represents each position as the sum of learned embeddings for its board, row, and column, whereas \texttt{FLAT} assigns a learned embedding to each absolute sequence position. 
Combining these choices yields four variants: \texttt{GRID}+\texttt{PAIR}, \texttt{FLAT}+\texttt{PAIR}, \texttt{GRID}+\texttt{BOARD}, and \texttt{FLAT}+\texttt{BOARD}.

\begin{table}[ht]
\caption{Development-set solve rates (\%) for the four input encoding variants across the seven models. A dagger indicates that the run had not fully converged by the end of training.}
\label{tab:encoding_ladder}
\centering
\small
\setlength{\tabcolsep}{8pt}
\begin{tabular}{@{}ccccc@{}}
\toprule
Model ID
& \texttt{GRID+PAIR}
& \texttt{FLAT+PAIR}
& \texttt{GRID+BOARD}
& \texttt{FLAT+BOARD} \\
\midrule
1 & 32.93 & \textbf{36.11} & 20.78$^{\dagger}$ & 33.22 \\
2 & \textbf{45.89} & 44.63 & 41.93 & 45.63 \\
3 & 79.70 & \textbf{82.70} & 79.78 & 81.74 \\
4 & \textbf{89.33} & 89.04 & 87.44 & 88.22 \\
5 & 91.85 & 92.26 & 91.96 & \textbf{95.00} \\
6 & 96.70 & 96.48 & \textbf{97.19} & 44.96$^{\dagger}$ \\
7 & \textbf{97.70} & 97.59 & 97.37 & 80.22$^{\dagger}$ \\
\bottomrule
\end{tabular}
\end{table}

\paragraph{Evaluation protocol.}
We evaluate all four variants using each of the seven model architectures in Appendix~\ref{appendix:model}, resulting in $28$ runs.
All runs use the same $30$M training instances, eight training epochs, a batch size of $512$, and a random seed of $42$.
We use AdamW with a peak learning rate of $1.5\times10^{-4}$, $1{,}000$ warmup steps, weight decay of $0.01$, gradient clipping at $1.0$, and dropout of $0.1$.
Table~\ref{tab:encoding_ladder} reports greedy solve rates on the held-out development set after the final epoch.

\paragraph{Encoding selection.}
For model ID~7, the three converged variants differ by only $0.33$ percentage points, indicating that their final solve rates are effectively comparable.
The \texttt{PAIR} encoding nevertheless provides a shorter input sequence and more stable optimization: all $14$ \texttt{PAIR} runs converged within the training budget, whereas three \texttt{BOARD} runs did not.
Within the \texttt{PAIR} encoding, \texttt{GRID} and \texttt{FLAT} achieve nearly identical solve rates at the largest model sizes. 
Nevertheless, across all seven model sizes, \texttt{GRID}+\texttt{PAIR} achieves the highest solve rate in three cases, more than any other variant, and is therefore selected for the main experiments.

\section{Training and Development Sets}
\label{appendix:train_gen}

The training set contains 30M instances with reference sequence lengths $L\in\{1,\ldots,9\}$.
For each instance, we sample $L$ using the weights in Table~\ref{tab:train_lengths}, and then sample $L$ actions independently from the ten atomic action types, with their arguments randomly sampled from valid locations. 
Sequences whose combined effect is the identity transformation are resampled. 
Each accepted sequence is executed on an initial board to obtain the target board. 
If the target board is identical to the initial board, the initial board is resampled. 
These checks exclude instances that can be solved without performing any action, ensuring that every training instance requires at least one action. 

\begin{table}[ht]
\caption{Sampling weights, counts, and actual shares of the 30M training instances by reference sequence length $L$. Actual shares differ from the weights by at most 0.015 points.}
\label{tab:train_lengths}
\centering
\small
\setlength{\tabcolsep}{10pt}
\begin{tabular}{cccc}
\toprule
Reference length $L$ & Sampling weight (\%) & Training instances & Actual share (\%) \\
\midrule
1 & 4.000  & 1,197,610 & 3.992 \\
2 & 4.000  & 1,195,963 & 3.987 \\
3 & 5.000  & 1,498,398 & 4.995 \\
4 & 6.000  & 1,797,360 & 5.991 \\
5 & 9.000  & 2,699,700 & 8.999 \\
6 & 13.000 & 3,904,518 & 13.015 \\
7 & 17.000 & 5,101,514 & 17.005 \\
8 & 20.000 & 6,002,666 & 20.009 \\
9 & 22.000 & 6,602,271 & 22.008 \\
\midrule
Total & 100.000 & 30,000,000 & 100.000 \\
\bottomrule
\end{tabular}
\end{table}

Beyond these validity checks, we use a non-uniform length distribution to help the models learn the task more effectively. 
Preliminary experiments showed that using the same number of instances at each length led to lower solve rates at lengths 7--9, while increasing the proportion of longer sequences improved performance at these lengths and maintained performance at shorter lengths. 
We therefore generate the training set using the sampling weights reported in Table~\ref{tab:train_lengths}.

The development set contains 2,700 instances, with 300 instances sampled for each reference sequence length $L\in\{1,\ldots,9\}$. 
Unlike the training set, the development set is balanced across reference sequence lengths. 
Duplicate instances with the same initial and target boards are removed before splitting, and the training and development sets are verified to contain no shared instance.

\end{document}